\documentclass[10pt,twocolumn,letterpaper]{article}

\usepackage{cvpr}              

\usepackage[dvipsnames]{xcolor}

\usepackage{graphicx}
\usepackage{amsmath}
\usepackage{amssymb}
\usepackage{booktabs}
\usepackage [switch]{lineno}
\usepackage{algorithm}
\usepackage{algorithmic}
\usepackage{booktabs}
\usepackage{multirow}
\usepackage{epsfig}
\usepackage{newfloat}
\usepackage{bm}
\usepackage{colortbl}
\usepackage{color}
\usepackage{pifont}
\usepackage{makecell} 
\usepackage{wrapfig}
\usepackage{xspace}
\usepackage{hhline}
\usepackage[accsupp]{axessibility}

\definecolor{cvprblue}{rgb}{0.21,0.49,0.74}
\usepackage[pagebackref,breaklinks,colorlinks,citecolor=cvprblue]{hyperref}

\def\paperID{936} 
\def\confName{CVPR}
\def\confYear{2024}

\title{Test-Time Domain Generalization for Face Anti-Spoofing}

\author{Qianyu Zhou$^{1}$\footnotemark[1], 
Ke-Yue Zhang$^{2}$\thanks{\textit{Equal contribution. }}, 
Taiping Yao$^2$, 
Xuequan Lu$^3$, Shouhong Ding$^2$\footnotemark[2],
Lizhuang Ma$^1$\thanks{\textit{Corresponding author.}}
\vspace{1mm}
\\$^1$Shanghai Jiao Tong University; $^2$Youtu Lab, Tencent; $^3$
La Trobe University.\\
$^1${\tt\small zhouqianyu@sjtu.edu.cn}, $^1${\tt\small ma-lz@cs.sjtu.edu.cn},  \\ $^2${\tt\small
\{zkyezhang,taipingyao,ericshding\}@tencent.com}, $^3${\tt\small b.lu@latrobe.edu.au }\\
}

\begin{document}
\maketitle
\begin{abstract}
Face Anti-Spoofing (FAS) is pivotal in safeguarding facial recognition systems against presentation attacks. While domain generalization (DG) methods have been developed to enhance FAS performance, they predominantly focus on learning domain-invariant features during training, which may not guarantee generalizability to unseen data that differs largely from the source distributions. Our insight is that testing data can serve as a valuable resource to enhance the generalizability beyond mere evaluation for DG FAS.  In this paper, we introduce a novel Test-Time Domain Generalization (TTDG) framework for FAS, which leverages the testing data to boost the model's generalizability. Our method, consisting of Test-Time Style Projection (TTSP) and Diverse Style Shifts Simulation (DSSS), effectively projects the unseen data to the seen domain space. In particular, we first introduce the innovative TTSP to project the styles of the arbitrarily unseen samples of the testing distribution to the known source space of the training distributions. We then design the efficient DSSS to synthesize diverse style shifts via learnable style bases with two specifically designed losses in a hyperspherical feature space. Our method eliminates the need for model updates at the test time and can be seamlessly integrated into not only the CNN but also ViT backbones. Comprehensive experiments on widely used cross-domain FAS benchmarks demonstrate our method's state-of-the-art performance and effectiveness. 
\end{abstract}    
\section{Introduction}

\begin{figure}
\includegraphics[width=\linewidth]{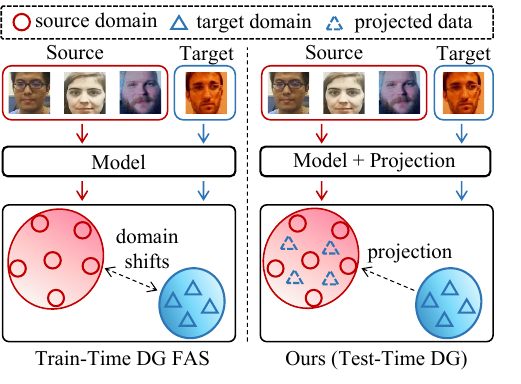}
\vspace{-7mm}
\caption{Conventional DG FAS approaches typically learn domain-invariant features at train time but cannot guarantee generalizability to unseen data that largely differ from source domains. 
In contrast, we propose test-time DG for FAS that projects the unseen testing data to the seen space, thus
enhancing the generalizability of FAS model without any model updates at test time. }
\vspace{-5mm}
\label{fig:illustrator}
\end{figure}

Face anti-spoofing (FAS) is critical in safeguarding face recognition~\cite{deng2019arcface,kemelmacher2016megaface,li2021spherical,wang2021facex} systems against different types of presentation attacks, such as printed photos or replaying videos. 
To address these presentation attacks, researchers have developed a range of FAS approaches, including those based on handcrafted features~\cite{maatta2011face,LBP01,2014Context,HoG01,2015Face,2017Face,2014dynamic}, as well as methods relying on deep learning for feature extraction~\cite{DeepBinary00,yu2021revisiting,zhang2021structure,lin2019face,disentangle01,liu2021face,cai2022learning,wang2022learning,wang2023consistency}. 
While these techniques have shown promising results within specific datasets, they often struggle to perform well when confronted with unseen domains due to the distribution shifts.

To improve the performance in unseen environments, recent research has introduced domain generalization (DG) techniques into FAS tasks. 
Some adversarial learning~\cite{shao2019multi,jia2020single,wang2022domain} techniques tend to align the domain distributions via mini-maxing the domain discriminator, and meta-learning~\cite{liu2021dual,liu2021adaptive,chen2021generalizable,zhou2022adaptive,du2022energy,jia2021dual} methods tend to simulate the unseen domain from the source domains. 
Other methods, \emph{e.g.,} instance whitening~\cite{zhou2023instance} and contrastive learning~\cite{wang2022domain}, align various instances in a self-supervised manner. 
However, all these methods focus on learning domain-invariant features during training to enhance generalization. 
As a result, they may still encounter performance degradation when dealing with unseen domains that have a significantly large discrepancy with the source domains.

To mitigate this issue, some domain adaptation~\cite{kim2023style,kim2022style,zhou2022generative,wang2021self} based FAS approaches aim to directly leverage the target data to align the testing distributions with the training ones.
Nevertheless, such methods suffer from two limitations in utilizing the target data.
Firstly, they necessitate updating the models for the target domain, which imposes a significant computational burden and in turn, severely impacts the performance in the source domains.
Secondly, they require a large amount of testing data for adaptation, which is not always available in realistic scenarios.

A natural question is can we leverage the target data in a more effective manner to enhance the generalizability of the FAS model?
To address this question, we propose a novel Test-Time Domain Generalization (TTDG) framework for DG FAS. 
Unlike traditional DG FAS methods which solely focus on training data, our insight is that testing data can serve as a valuable resource to enhance the generalizability beyond mere prediction before classification. 
Our TTDG framework elegantly utilizes the testing data to improve the performance of the FAS model without any model updates at test time, as shown in Figure~\ref{fig:illustrator}. 
Firstly, an innovative Test-Time Style Projection (TTSP) is introduced to dynamically leverage unseen samples by projecting them to the known source space based on the similarity between the unseen samples and the training distributions. 
Specifically, to accurately model the training distributions, we are motivated to design a series of style bases to handle the various domain shifts in the training data, \emph{e.g.,} illumination and color, \emph{etc.}
However, manually selecting style bases for source domains is cumbersome and time-expensive, and there is no guarantee that selected bases will fully capture the domain shifts in the training data, and accurately project the unseen samples into the correct position during the test time. 
As such, we design the efficient Diverse Style Shifts Simulation (DSSS) with two new losses to model diverse style shifts via learnable style bases in a hyperspherical feature space. 
The first loss is a style diversity loss that encourages each learnable style basis to be orthogonal in the hyperspherical space, thus increasing the diversity in the style bases. 
The second one is a content consistency loss that ensures each projected feature is closely aligned with its corresponding content feature, preventing content distortion. 
Our TTDG is model-agnostic and can be seamlessly integrated into 
not only CNN but also ViT backbones.

Our contributions are three-fold:

$\bullet$  We offer a new perspective for DG FAS that leverages the testing data to enhance the generalizability beyond evaluation and propose a novel Test-Time Domain Generalization (TTDG) framework for FAS. To the best of our knowledge, this is the first work that studies test-time DG for FAS. 

$\bullet$ We present Test-Time Style Projection (TTSP) to project the unseen samples to the seen source distributions via the aggregation of a set of style bases. Besides, we design Diverse Style Shifts Simulation (DSSS) with two new losses to synthesize diverse distribution shifts via learnable style bases in a hyperspherical feature space.

$\bullet$ We conduct extensive experiments that demonstrate the state-of-the-art performance and effectiveness of our TTDG on widely used cross-domain FAS benchmarks. 
\section{Related Work}
\noindent \textbf{Face Anti-Spoofing.} 
Face anti-spoofing (FAS) aims to determine whether an image captures a genuine human face or a presentation attack, such as a printed photo or video replay. 
Early FAS research relied on hand-crafted features \cite{2016Secure,boulkenafet2015face,maatta2011face,LBP01,2014Context,HoG01} to detect spoof patterns.
With the rise of deep learning, various techniques, \emph{e.g.,} classification-based methods~\cite{DeepBinary00,DeepBinary01,DeepBinary02,2014Learn,wang2022patchnet,liu2019deep,hu2022structure}, regression-based methods~\cite{atoum2017face,kim2019basn,2018Learning,BCN,yu2021revisiting,CDCN,yu2020fas,wu2020single,disentangle01,qin2021meta,wang2020deep}, and generative models~\cite{jourabloo2018face,liu2022spoof,STCN,liu2021face,wu2021dual} have been explored to enhance FAS performance.
Recently, vision Transformer~\cite{dosovitskiy2020image,touvron2021training,mehta2021mobilevit} has shown promising potential in FAS~\cite{george2021effectiveness,wang2022learning,liao2023domain,huang2023ldcformer,liu2023fm,wang2022face,liu2023ma,hong2023domain}. 
Despite their gratifying progress in the intra-dataset settings, their performances degrade significantly when applied to different target domains.
To mitigate this challenge, domain adaptation techniques~\cite{ganin2015unsupervised,tzeng2017adversarial,zhou2023self,zhou2022context,zhou2020uncertainty,zhou2022domain,PIT,song2024simada,song2024ba,liu2024cloud,xu2021semi,guo2021label,feng2020dmt} have been recently integrated into FAS~\cite{panwar2021unsupervised,wang2021vlad,DR-UDA,2018Unsupervised,jia2021unified,wang2021rgb,liu2022source,liu2024source}, but the target data is not always accessible in real scenarios and might fail these methods. Thus, domain generalization techniques~\cite{li2018domain,li2018learning,long2023rethink,long2023diverse}
have been introduced to improve the performance on unseen domains via adversarial learning~\cite{shao2019multi,jia2020single,wang2022domain,kwak2023liveness}, meta-learning ~\cite{liu2021dual,liu2021adaptive,chen2021generalizable,zhou2022adaptive,du2022energy,jia2021dual}, instance whitening~\cite{zhou2023instance} and \emph{etc}~\cite{hu2024rethinking,hu2024domain,zheng2023learning,liu2023towards,liu2022causal}. Nevertheless, almost all of them merely focus on learning domain-invariant features at train time and may fail in real-world scenarios that differ significantly from the source domains. Besides, they overlook the role of testing data beyond just evaluation. 
In this work, we propose a novel perspective that leverages the testing data to boost the generalizability of FAS models.

\begin{figure*}
\includegraphics[width=1.0\textwidth]{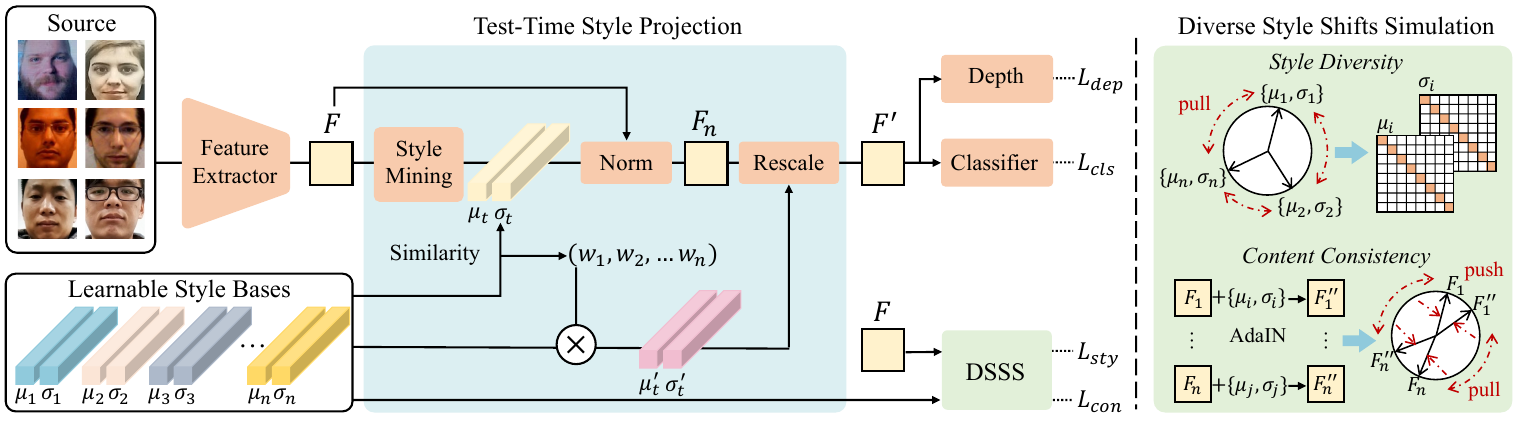}
\vspace{-6mm}
\caption{
Overview of the proposed Test-Time Domain Generalization (TTDG) framework for DG FAS. 
In particular, we first introduce Test-Time Style Projection (TTSP) to project arbitrarily unseen samples to the known source space based on the similarity between the unseen sample and the style bases. 
We then design Diverse Style Shifts Simulation (DSSS) to synthesize diverse style shifts via learnable style bases. $\mathcal{L}_{\text{sty}}$ and $\mathcal{L}_{\text{con}}$ are two new losses for maximizing the style diversity and content consistency in a hyperspherical feature space. 
Our TTDG eliminates the need for model updates at test time and can be seamlessly integrated into
the CNN and ViT backbones.
}
\vspace{-2mm}
\label{fig:framework}
\end{figure*}

\noindent \textbf{Test-Time Domain Adaptation and Domain Generalization.} Test-time adaptation (TTA)~\cite{wang2020tent,chen2022contrastive,wang2022continual,niu2022efficient,zhang2022memo} has been studied to enhance the model's transferability to the target domain, where the classifier
is updated partially or fully using incoming batches of test samples. Kim \emph{et al.} \cite{kim2023style,kim2022style} proposed a style selective normalization for test-time adaptive FAS. Similarly, \cite{zhou2022generative} and \cite{wang2021self} aim to align the target domain with the source ones in a reverse manner. 
Nevertheless, they need batches of the target data
for gradient updates~\cite{kim2023style,kim2022style,wang2021self} or an additional model for fine-tuning~\cite{zhou2022generative}, which requires considerable computation burden during the test time.
In addition, acquiring such a large number of testing data is impractical in realistic scenarios.
In contrast to these methods, test-time domain generalization (TTDG) is more challenging since it does not require any model updates at test time. Park \emph{et al.} introduced test-time style shifting~\cite{park2023test}, which shifts the style of the test sample to the nearest source domain before making the prediction. Besides, \cite{ijcai2022p240,jiang2022domain,huang2023style} pulled the target closer to the source distributions via Fourier transformation and normalization, \emph{etc.} 
Despite their encouraging advancements, they suffer from two limitations. Firstly, style bases in \cite{ijcai2022p240,park2023test,huang2023style} are roughly defined and inadaptive, either one for each domain or one for all domains, which is inapplicable to FAS tasks that typically have a mixture of domains. In contrast, we do not require any domain label and design a set of learnable style bases that are more fine-grained to automatically capture style shifts. Secondly, their projections lack a clear objective for optimization, making projection less reliable for FAS. Conversely, 
we propose explicit optimization goals to ensure the reliability of projection.

\section{Methodology}
Figure~\ref{fig:framework} shows the overview of the proposed Test-Time Domain Generalization (TTDG) framework, which aims to leverage the testing data to improve the generalizability of FAS models. Our TTDG framework consists of two key components.
Firstly, we present Test-Time Style Projection (TTSS) to project the styles of the unseen samples to the style representation space built on style bases, according to the similarity between the unseen style and style bases. In addition, we design Diverse Style Shifts Simulation (DSSS) with two new losses to synthesize diverse distribution shifts via learnable style bases in a hyperspherical feature space.

\subsection{Theoretical Analysis}
To perform domain generalization, we need to first understand the distribution shifts measured by $\mathcal{H}$-divergence~\cite{ben2010theory}: 
\begin{small}
\begin{equation}
\begin{gathered}
    d_{\mathcal{H}}(\mathcal{D}_{s}, \mathcal{D}_{t})= 2 \sup _{h \in \mathcal{H}}\left|\operatorname{Pr}_{x\sim{\mathcal{D}_{s}}}[h(x)=1]
    -\operatorname{Pr}_{x\sim{\mathcal{D}_{t}}}[h(x)=1]\right|
\end{gathered}
    \label{dis-shift}
\end{equation}
\end{small}
where classifier $h: \mathcal{X} \rightarrow\{0,1\}$.
Then, \cite{albuquerque2019generalizing} defines the convex hull $\Lambda_{s}$ of $\mathcal{D}_{s}$ that is a set of mixture of source domains:
\begin{equation}
\Lambda_{s}=\left\{\sum_{i=1}^{K} \eta_{i} \mathcal{D}_{s}^{i} \mid \eta \in \Delta_{K-1}\right\},
\end{equation}
where  $\eta$ denotes non-negative coefficient in the $(K-1)$-dimensional simplex $\Delta_{K-1}$. Next, an ideal case $\Bar{\mathcal{D}}_{t} \in \Lambda_{s}$ is assumed that the ideal target domain $\Bar{\mathcal{D}}_{t}$ lies in the source domain convex hull $\Lambda_{s}$. Under this assumption, the risk $\epsilon_{t}(h)$ on the target domain $\mathcal{D}_{t}$ is upper-bounded \cite{albuquerque2019generalizing} by:
\begin{equation}
\epsilon_{t}(h) \leq  \sum_{i=1}^{K} \eta_{i} \epsilon^{i}_{s}(h) + \gamma + \zeta,
\label{eq:3}
\end{equation}
where, on the right side, the first term represents the risks over all source domains, and the second term $\gamma=d_{\mathcal{H} \Delta \mathcal{H}}(\Bar{\mathcal{D}}_{t}, D_{t})$ denotes the $\mathcal{H}$-divergence between the ideal target $\Bar{\mathcal{D}}_{t}$ and the real target domain $D_{t}$, and the third term $\zeta=\sup _{D_{s}^{\prime}, D_{s}^{\prime \prime} \in \Lambda_{s}} d_{\mathcal{H} \Delta \mathcal{H}}\left(D_{s}^{\prime}, D_{s}^{\prime \prime}\right)$ is the largest $\mathcal{H}$-divergence between any pair of source domains. $\mathcal{H} \Delta \mathcal{H}$ corresponds to $\left\{h(x) \oplus h^{\prime}(x) \mid h, h^{\prime} \in \mathcal{H}\right\}$. The first term can be minimized by empirical risk minimization (ERM), and the second term is hard to minimize due to no access to the target domain at training, and the third term can be minimized by removing the source domain-specific information which is the style information in the context of this work. 

Almost all previous DG FAS approaches~\cite{shao2019multi,jia2020single,wang2022domain,liu2021dual,liu2021adaptive,chen2021generalizable,zhou2022adaptive,du2022energy,jia2021dual,le2024gradient,lin2024suppress} assume that the ideal target domain $\Bar{\mathcal{D}}_{t}$ is covered by the source convex hull $\Lambda_{s}$, and thus a model can achieve acceptable performance on the target domain by just minimizing the source divergence (Eq.~\eqref{eq:3}).
However, this assumption typically does not hold in reality since 
the realistic target data may differ significantly from the source domains. 
Recent works~\cite{wang2022domain,zhou2023instance} rely on data augmentation to generate the data outside the source distributions, possibly extending the source convex hull $\Lambda_{s}$ (\emph{i.e.,} $\gamma \rightarrow 0$). Nevertheless, the augmented source domain might not fully overlap with the target domain, leading to the failure of generalization of existing models on unseen domains. 



The above analysis motivates us to re-think the domain generalization for FAS. 
Our core idea is to leverage the testing data as a valuable resource to enhance the generalizability. Our TTSP (Section \ref{sec:ttdp}) and our DSSS (Section \ref{sec:dsss}) aim to pull the target data closer to the source convex hull $\Lambda_{s}$,  thus reducing the difference between $D_{t}$ and $\Bar{D_{t}}$ ($\gamma \rightarrow 0$). 



\vspace{-2mm}

\subsection{Test-Time Style Projection}
\label{sec:ttdp}
Our test-time style projection (TTSP) strategy aims to project the styles of the
unseen test samples to the known space to handle arbitrary unseen domains during the testing phase. To achieve this goal, there are two key questions that we need to explore for DG FAS.
Firstly, how to represent the known style space to the utmost extent? Secondly, how to effectively shift or project the unseen sample to the known domains? We address these questions below. 

Regarding the first question, our idea is to build a robust style representation space that can be defined by a series of style bases since various presentation attacks primarily vary in terms of styles, such as illumination, color, \emph{etc.}, and such style differences are the main factor in leading to domain shifts. Therefore, the key to improving the generalizability 
of the FAS model lies in narrowing the style gaps. Previous FAS studies~\cite{wang2022domain,kim2023style, kim2022style,zhou2022generative,zhou2023instance} have demonstrated that the statistics of the latent features of FAS models can reflect the style information of the input image $x_t$, and most of them commonly employ the channel-wise mean and variance of these features to represent the style distribution of $x_t$.
Following them,  $F_t \in \mathbb{R}^{C \times H \times W}$ is denoted as the feature of $x_t$ from the feature extractor, where $C$ denotes the number of channels. The channel-wise mean $\mu_t(F_{t}) \in \mathbb{R}^C$ and variance $\sigma_t(F_{t}) \in \mathbb{R}^C$ of the feature $F_t$ can be calculated as follows (the Style Mining part in Figure~\ref{fig:framework}):
\begin{small}
\begin{equation}
\mu_t  =\frac{1}{H W} \sum_{h=1}^{H} \sum_{w=1}^{W} F_t, 
\sigma_t  =\sqrt{\frac{1}{H W} \sum_{h=1}^{H} \sum_{w=1}^{W}\left(F_t -\mu_t \right)^2},
\end{equation}
\end{small}

We design a series of style bases $B_{sty}=\left\{\left(\mu_b^n, \sigma_b^n \right)\right\}_{n=1}^N$ to preserve the style information of source domains, where $N$ denotes the number of style bases. Then, we build a style representation space based on these style bases $B_{sty}$ for realizing the test-time projection of the unseen style. The way of properly selecting these style bases for the DG FAS task will be discussed in Section~\ref{sec:dsss}. 

As for the second question, we aim to project the style of unseen faces into the style representation space as a weighted combination of style bases. To achieve this goal, firstly, we calculate the cosine distance to estimate the style distribution discrepancy $d_n$  between the current image $x_t$ and the $n$-th style basis $\left(\mu_b^n, \sigma_b^n \right)$, defined as follows: 
\begin{equation}
d_n = \frac{\mu_t \cdot \mu_b^n }{\| \mu_t \| \cdot \| \mu_b^n \|} +  \frac{\sigma_t \cdot \mu_b^n }{\| \sigma_t \| \cdot \| \mu_b^n \|}, w_n=\frac{ e^{d_n} }{\sum_{n=1}^N e^{d_n}},
\label{eq:cosine_sim}
\end{equation}
where $w_n$ denotes the estimated weighting factor calculated by the Softmax operation such that the sum of $w =\left\{w_n \mid n=1,2, \ldots, N\right\}$ is equal to 1. 

Next, we can obtain the projected style $\left(\mu_{t}^{\prime}, \sigma_{t }^{\prime}\right)$ by the weighted combination of style bases as follows:
\begin{equation}
\mu_{t}^{\prime}=\sum_{n=1}^N w_n \cdot \mu_b^n, \qquad  \sigma_{t }^{\prime}=\sum_{n=1}^N w_n \cdot \sigma_b^n,
\end{equation}
With the projected styles $\left(\mu_{t}^{\prime}, \sigma_{t }^{\prime}\right)$ and input feature $F_t$ of the $t$-th sample, the style projected feature $F_{t}^{\prime}$ is defined as:
\begin{equation}
    F_{t}^{\prime} = \sigma_{t}^{\prime}(F_{t}) \cdot \left(\frac{F_{t}-\mu_t(F_{t})}{\sigma_t(F_{t})}\right)+\mu_{t}^{\prime}(F_{t}),
    \label{eq:adain}
\end{equation}
As such, each test sample that has a style gap with the
source domains will be projected to the source domains via the aggregation of the set of style bases. For example, when the unseen faces have a large discrepancy with the style representation space, the nearest style basis that the model is familiar with will have a large contribution to the projection, and the farthest style basis will have less contribution. 

\subsection{Diverse Style Shifts Simulation}
\label{sec:dsss}
Although building a style representation space is promising for test-time DG, manually selecting style bases of source domains is cumbersome and time-expensive, especially when the style space is continually changing during the model updating, and it requires to re-select the bases from all the source domains in every epoch. This manner will complicate the procedure and largely reduce the model's efficiency. Besides, there is no guarantee that manually selected style bases will represent the style representation space to the utmost extent. For example, those selected bases might merely include dominant styles that have high frequency and ignore the rare styles with low frequency in the source domains. As a result, these selected style bases may steer the model in the incorrect direction at test time. 

To address these issues, instead of utilizing the manually selected style bases, we propose a novel strategy that uses learnable style bases in hyperspherical feature space to synthesize diverse style shifts for FAS. Our method, namely Diverse Style Shifts Simulation (DSSS), is more efficient and effective. In addition, two new loss functions are specifically introduced to guide the learning of learnable style bases. We will describe them in detail below. 

\noindent \textbf{Style Diversity Loss.} To maximize the diversity of $N$ style bases in a hyperspherical feature space, we present a style diversity loss such that the $i$-th style basis $B_{sty}^i = \left(\mu_b^i, \sigma_b^i \right)$ is orthogonal to other ones $B_{sty}^k \in \left\{\left(\mu_b^n, \sigma_b^n \right)\right\}_{k=1, k!=i}^N$. Regarding this, the style diversity loss $\mathcal{L}_{\text {style }}$ for learning the $i$-th style basis is computed by:
\begin{equation}
\mathcal{L}_{\text{sty}} = \sum_{\substack{k=1 \\ k \neq i}}^{N}
 \left| \frac{\mu_b^i}{\left\| \mu_b^i \right\|} \cdot \frac{\mu_b^k}{\left\| \mu_b^k \right\|} \right| + \sum_{\substack{k=1 \\ k \neq i}}^{N}
 \left| \frac{\sigma_b^i}{\left\| \sigma_b^i \right\|} \cdot \frac{\sigma_b^k}{\left\| \sigma_b^k \right\|} \right|,
\end{equation}
The objective of the style loss $\mathcal{L}_{\text {style }}$ is to minimize the absolute value of the cosine similarity between the $i$-th style basis and every other existing style basis. When this loss value reaches zero, it signifies that the $i$-th style basis has achieved orthogonality with respect to all the other ones.

\begin{table*}[t!]
\centering
\begin{center}
\resizebox{1.0\textwidth}{!}{%
\begin{tabular}{r  c c  c c  c c  c c }
\toprule
\multirow{2}{*}{\textbf{Methods}} &
\multicolumn{2}{c}{\textbf{I\&C\&M to O}} &
\multicolumn{2}{c}{\textbf{O\&C\&M to I}} &
\multicolumn{2}{c}{\textbf{O\&C\&I to M}} &
\multicolumn{2}{c}{\textbf{O\&M\&I to C}} \\
&HTER(\%) &AUC(\%) &HTER(\%) &AUC(\%) &HTER(\%) &AUC(\%) &HTER(\%) &AUC(\%)\\
\midrule
MADDG~\cite{shao2019multi}&$27.98$ &$80.02$  &$22.19$ &$84.99$  &$17.69$ &$88.06$ &24.50 &84.51\\
D$^2$AM~\cite{chen2021generalizable} &15.27 &90.87  &15.43 &91.22  &12.70 &95.66 &20.98 &85.58\\
SSDG~\cite{jia2020single}& 25.17 & 81.83 & 18.21 &94.61  &16.67 &90.47   & 23.11 & 85.45 \\
RFM~\cite{shao2020regularized}&16.45 &91.16  &17.30 &90.48 &13.89 &93.98 &20.27 &88.16\\
DRDG~\cite{liu2021dual}&15.63 & 91.75  & 15.56 & 91.79 & 12.43 & 95.81  & 19.05 & 88.79\\
ANRL~\cite{liu2021adaptive}& 15.67 & 91.90   & 16.03 & 91.04 & 10.83 & 96.75  & 17.85 & 89.26\\
FGHV~\cite{liu2022feature} &13.58  &93.55    &16.29  &90.11  &9.17  &96.92   &12.47  &93.47 \\
SSAN~\cite{wang2022domain} &19.51  &88.17  &14.00 &94.58  & 10.42 &94.76 & 16.47 &90.81 \\
AMEL~\cite{zhou2022adaptive} & 11.31 & 93.96  &18.60 &88.79 &10.23 &96.62 &11.88 &94.39\\
EBDG~\cite{du2022energy} & 15.66 & 92.02  &18.69 &92.28 &9.56 &\textbf{97.17} &18.34 &90.01\\
IADG~\cite{zhou2023instance} & 11.45 & 94.50 & 11.04 & 93.15 & 8.45 & 96.99 & 12.74 &  94.00 \\
\!\cellcolor{gray!9.0} Ours (TTDG) \!
& \!\cellcolor{gray!9.0} \textbf{10.00} \! & \!\cellcolor{gray!9.0} \textbf{95.70} \! &\!\cellcolor{gray!9.0} \textbf{6.50} \!& \!\cellcolor{gray!9.0} \textbf{97.98} \! &\!\cellcolor{gray!9.0} \textbf{7.91} \!& \!\cellcolor{gray!9.0} 96.83 \!& \!\cellcolor{gray!9.0} \textbf{8.14} \!& \!\cellcolor{gray!9.0} \textbf{96.49} \!\\
\midrule
ViTranZFAS~\cite{george2021effectiveness}  & 15.67 & 89.59 & 16.64 & 85.07 &10.95  &95.05 & 14.33 & 92.10\\
TTN-S~\cite{wang2022learning} & 12.64 & 94.20 & 14.15 & 94.06 & 9.58 & 95.79 & 9.81 & 95.07\\
DiVT-V~\cite{liao2023domain} & 18.06  & 90.21 & \textbf{5.71 }& 97.73 & 10.00 & 96.64 & 14.67 & 93.08  \\
\!\cellcolor{gray!9.0} Ours (TTDG-V) \!
& \!\cellcolor{gray!9.0} \textbf{10.00} \! & \!\cellcolor{gray!9.0} \textbf{96.15} \! &\!\cellcolor{gray!9.0} 9.62 \!& \!\cellcolor{gray!9.0} \textbf{98.18} \! &\!\cellcolor{gray!9.0} \textbf{4.16} \!& \!\cellcolor{gray!9.0} \textbf{98.48} \!& \!\cellcolor{gray!9.0} \textbf{7.59} \!& \!\cellcolor{gray!9.0} \textbf{98.18} \!\\
\bottomrule
\end{tabular}}
\end{center}
\vspace{-6mm}
\caption{Comparison with the state-of-art FAS methods on four testing domains. TTDG-V denotes TTDG with ViT-Base~\cite{dosovitskiy2020image} backbone. }
\label{tab:DG_SOTA_3to1}
\vspace{-2.5mm}
\end{table*}

\noindent \textbf{Content Consistency Loss.}
Merely using style diversity loss to learn style bases might potentially result in a less desirable outcome because learnable style bases could substantially distort the content information when used to generate a style-content reassembled feature. Thus, for each basis, we encourage the style-content feature to exhibit the highest consistency with its corresponding content feature.

Specifically, for each input feature $F_{t}$, we randomly select a style basis $B_{sty}^i$ from the style bases set, and reassemble a style-content feature $F_{t}^{\prime\prime}$ using Eq.~(\ref{eq:adain}). Then, we devise a content consistency loss $\mathcal{L}_{\text {content }}$ that maximizes the cosine similarity scores between $F_{t}^{\prime\prime}$ and $F_{t}$ as follows:

\begin{equation}
z_{mt}=\frac{ F_{t}^{\prime\prime} }{\left\| F_{t}^{\prime\prime} \right\|_2} \cdot \frac{ F_{m} }{\left\|  F_{m} \right\|_2} ,
\end{equation}
\begin{equation}
    \mathcal{L}_{\text {con}}=-\frac{1}{M} \sum_{t=1}^M \log \left(\frac{\exp \left(z_{tt}\right)}{\sum_{m=1}^M \exp \left(z_{mt}\right)}\right),
\end{equation}
where $M$ denotes the batch size and $z_{mt}$ is the cosine similarity score between the style-content reassembled feature $F_{t}^{\prime\prime}$ and the content feature $F_{m}$ of the $m$-th sample. This content loss $\mathcal{L}_{\text {content }}$ encourages each style-content feature to be closer to its corresponding original feature. This way forces each $i$-th style basis $B_{sty}^i$ to preserve content information when used to synthesize style-content features.

\subsection{Training and Inference}
\label{sec:loss}
\noindent \textbf{Training.} To ensure that the feature extractor captures task-relevant features $F_t$ of each sample $X_t$ for good classification, we introduce a binary classification loss $\mathcal{L}_{\mathrm{cls}}$:

\begin{equation}
\begin{aligned}
\mathcal{L}_{\mathrm{cls}} = -\sum_{(X_t,Y_t^{cls})}Y_t^{cls} \log(\text {Cls}(F_t)),
\end{aligned}
\end{equation}
In this equation, $\text {Cls}$ represents the binary classifier responsible for distinguishing genuine faces from face presentation attacks. Here, $X_t$ corresponds to the input image, and $Y_t^{cls}$ is the classification label, as illustrated in Figure~\ref{fig:framework}.

Previous research~\cite{2018Learning,liu2021adaptive,liu2021dual} has demonstrated the usefulness of depth information for guiding FAS at the pixel level. We follow them by utilizing a depth estimator ($\text {Dep}$), which estimates depth maps for live faces and zero maps for spoof faces. With the guidance of depth label $Y_t^{dep}$, we introduce the depth loss $\mathcal{L}_{Dep}$, defined as follows:
\begin{equation}
\begin{aligned}
\mathcal{L}_{\mathrm{dep}} = & \sum_{(X_t,Y_t^{dep})} \left| \text {Dep}(F_t) - Y_t^{dep} \right|_{2}^{2},
\end{aligned}
\end{equation}
Additionally, to ensure that our FAS model could well project the unseen sample to the known space during the test time, we simulate this projection process during the training, and the total training loss is defined as:
\begin{equation}
\begin{aligned}
    \mathcal{L}_{\mathrm{total}} = \mathcal{L}_{\mathrm{cls}} +  \lambda_d  \mathcal{L}_{\mathrm{dep}}  +  \mathcal{L}_{\mathrm{sty}} + \lambda_c  \mathcal{L}_{\mathrm{con}},
\end{aligned}
\end{equation}
Instead of manually re-selecting the style bases in each epoch, we jointly train the whole model with learnable style bases in every iteration, which is a more efficient manner.

\noindent \textbf{Inference.} During the test phase, unseen faces are fed into the feature extractor and then projected into the style representation space via our TTSP. The outputs are next fed into the classifier for making the final prediction. Note that different from existing TTA FAS methods~\cite{kim2023style,kim2022style,zhou2022generative,wang2021self}, \emph{our model does not require any parameter update at the test time, which is more flexible in real-world scenarios}. Moreover, \emph{our TTDG method can be seamlessly integrated into not only the CNN backbone but also the ViT backbone}.

\begin{table*}[t!]
\centering
\begin{center}
\resizebox{1.0\textwidth}{!}{%
\begin{tabular}{r  c c  c c  c c  c c }
\toprule
\multirow{2}{*}{\textbf{Methods}} &
\multicolumn{2}{c}{\textbf{I\&C\&M to O}} &
\multicolumn{2}{c}{\textbf{O\&C\&M to I}} &
\multicolumn{2}{c}{\textbf{O\&C\&I to M}} &
\multicolumn{2}{c}{\textbf{O\&M\&I to C}} \\
&HTER(\%) &AUC(\%) &HTER(\%) &AUC(\%) &HTER(\%) &AUC(\%) &HTER(\%) &AUC(\%)\\
\midrule
DCN~\cite{jiang2022domain} & 15.52& 90.44  & 18.75 & 87.23 & 14.16 & 95.19 & 15.74 & 91.51 \\
TF-Cal~\cite{ijcai2022p240} & 13.29 & 93.71 & 19.75 & 90.35 & 12.08 & 95.58 & 14.26 & 92.10 \\
Sty.-Pro~\cite{huang2023style} & 13.19 & 93.69 & 14.25 & 91.63 & 14.58 & 92.60 &14.81 & 92.63 \\
\!\cellcolor{gray!9.0} Ours (TTDG) \!
& \!\cellcolor{gray!9.0} \textbf{10.00} \! & \!\cellcolor{gray!9.0} \textbf{95.70} \! &\!\cellcolor{gray!9.0} \textbf{6.50} \!& \!\cellcolor{gray!9.0} \textbf{97.98} \! &\!\cellcolor{gray!9.0} \textbf{7.91} \!& \!\cellcolor{gray!9.0} \textbf{96.83} \!& \!\cellcolor{gray!9.0} \textbf{8.14} \!& \!\cellcolor{gray!9.0} \textbf{96.49} \!\\
\bottomrule
\end{tabular}}
\end{center}
\vspace{-6mm}
\caption{Comparison with test-time domain generalization methods. The bold numbers indicate the best performance.}
\label{tab:test_time_DG}
\vspace{-2.5mm}
\end{table*}

\begin{table}[t!]
\centering
\begin{center}
\resizebox{0.48\textwidth}{!}{%
\begin{tabular}{r  c c  c c }
\toprule
\multirow{2}{*}{\textbf{Methods}} &
\multicolumn{2}{c}{\textbf{M\&I to C}} &
\multicolumn{2}{c}{\textbf{M\&I to O}} \\
    &HTER(\%) &AUC(\%) &HTER(\%) &AUC(\%) \\
\midrule
MADDG~\cite{shao2019multi} & $41.02$ & $64.33$ &  $39.35$ & $65.10$ \\
SSDG~\cite{jia2020single} & $31.89$ & $71.29$ &  $36.01$ & $66.88$ \\
D$^2$AM~\cite{chen2021generalizable} &32.65 &72.04 &27.70 & 75.36 \\
DRDG~\cite{liu2021dual} &31.28 &71.50 &33.35 & 69.14 \\
ANRL~\cite{liu2021adaptive} & 31.06 & 72.12 & 30.73 &74.10 \\
SSAN~\cite{wang2022domain} & 30.00 & 76.20 & 29.44 & 76.62 \\
EBDG~\cite{du2022energy} & 27.97  & 75.84 & 25.94 & 78.28 \\
AMEL~\cite{zhou2022adaptive} &24.52 &82.12 & 19.68 & 87.01  \\
IADG~\cite{zhou2023instance} & 23.51 & 84.20 & 22.70 & 84.28 \\
\!\cellcolor{gray!9.0} Ours (TTDG)  \! & 
\!\cellcolor{gray!9.0} \textbf{17.77}  \! & \!\cellcolor{gray!9.0} \textbf{86.69} \! & \!\cellcolor{gray!9.0} \textbf{17.70}  \! & \!\cellcolor{gray!9.0} \textbf{90.09}   \! \\
\bottomrule
\end{tabular}
}
\end{center}
\vspace{-6mm}
\caption{Comparison results on limited source domains. }
\label{tab:DG_SOTA_2to1}
\vspace{-6mm}
\end{table}

\section{Experiments}
\subsection{Experimental Setup}
\label{sec:4.1}
\noindent \textbf{Datasets and Protocols.} 
We conducted experiments on four public FAS datasets, namely CASIA-MFSD (C)~\cite{Zhang2012A}, Idiap Replay-Attack (I)~\cite{2012Replay}, MSU-MFSD (M)~\cite{2015Face}, OULU-NPU (O)~\cite{2017OULU} to verify the efficacy of our approach. These datasets include print, paper cut, and replay attacks, and were gathered using different capturing devices, including diverse illumination conditions, various background scenes, and racial demographics. Thus, there exist substantial domain shifts among these datasets.
For all experiments, we strictly follow the same experimental protocols as previous DG FAS methods~\cite{shao2019multi,jia2020single,shao2020regularized,liu2021dual,liu2021adaptive,zhou2022adaptive,sun2023rethinking}.

\noindent \textbf{Implementation Details:} 
Following the precedent setting of previous DG FAS methods~\cite{shao2019multi,jia2020single,shao2020regularized,liu2021dual,liu2021adaptive}, we employ the same CNN~\cite{jia2020single,yu2021dual} and ViT-Base~\cite{dosovitskiy2020image} backbone to ensure fair comparisons.
During training, the hyperparameter $\lambda_c$ is empirically set to $0.4$, and $N$ is set to 64 for all experiments. Following prior works~\cite{liu2021adaptive,liu2021dual,zhou2022adaptive,zhou2023instance}, we utilize pseudo-depth maps generated by PRNet~\cite{2018Joint3D} for depth supervision and set $\lambda_d=0.1$ when training with the CNN-based backbone. As for training with ViT~\cite{dosovitskiy2020image}, we follow \cite{george2021effectiveness,wang2022learning,liao2023domain} and do not use the depth estimator ($\lambda_d=0$).
The Half Total Error Rate (HTER) and the Area Under Curve (AUC) are used as evaluation metrics. The lower HTER and higher AUC indicate better performance.

\begin{table*}[t!]
\centering
\begin{center}
\resizebox{1.0\textwidth}{!}{%
\begin{tabular}{c  c c  c c  c c  c c }
\toprule
\multirow{2}{*}{\textbf{Methods}} &
\multicolumn{2}{c}{\textbf{I\&C\&M to O}} &
\multicolumn{2}{c}{\textbf{O\&C\&M to I}} &
\multicolumn{2}{c}{\textbf{O\&C\&I to M}} &
\multicolumn{2}{c}{\textbf{O\&M\&I to C}} \\
&HTER(\%) &AUC(\%) &HTER(\%) &AUC(\%) &HTER(\%) &AUC(\%) &HTER(\%) &AUC(\%)\\
\midrule
TF-Cal~\cite{ijcai2022p240} & 12.74 & 93.50 & 14.00 & 93.18 & 13.75 & 91.06 & 13.33 & 93.36 \\
TTSS~\cite{park2023test} & 15.10 & 93.30 & 12.50 & 92.91 & 9.58 & 95.88 & 13.14 & 93.65 \\
Sty.-Pro~\cite{huang2023style} & 13.05 & 93.66 & 11.75 & 92.51 & 11.25 & 94.98 & 13.51 & 93.27 \\
\!\cellcolor{gray!9.0} Ours (TTSP) \!
& \!\cellcolor{gray!9.0} \textbf{10.00} \! & \!\cellcolor{gray!9.0} \textbf{95.70} \! &\!\cellcolor{gray!9.0} \textbf{6.50} \!& \!\cellcolor{gray!9.0} \textbf{97.98} \! &\!\cellcolor{gray!9.0} \textbf{7.91} \!& \!\cellcolor{gray!9.0} \textbf{96.83} \!& \!\cellcolor{gray!9.0} \textbf{8.14} \!& \!\cellcolor{gray!9.0} \textbf{96.49} \!\\
\bottomrule
\end{tabular}}
\end{center}
\vspace{-6mm}
\caption{Ablation studies on different test-time style shifting strategies on four testing domains.}
\label{tab:ablation_shifting}
\vspace{-2mm}
\end{table*}

\begin{table*}[t!]
\centering
\begin{center}
\vspace{-1mm}
\resizebox{1.0\textwidth}{!}{%
\begin{tabular}{c  c c  c c  c c  c c }
\toprule
\multirow{2}{*}{\textbf{Methods}} &
\multicolumn{2}{c}{\textbf{I\&C\&M to O}} &
\multicolumn{2}{c}{\textbf{O\&C\&M to I}} &
\multicolumn{2}{c}{\textbf{O\&C\&I to M}} &
\multicolumn{2}{c}{\textbf{O\&M\&I to C}} \\
&HTER(\%) &AUC(\%) &HTER(\%) &AUC(\%) &HTER(\%) &AUC(\%) &HTER(\%) &AUC(\%)\\
\midrule
Random Selection & 13.40 & 92.81 & 15.12 & 89.56 & 13.75 & 93.50 & 13.51 & 92.72 \\
FPS~\cite{qi2017pointnet++} Selection & 11.35 & 95.14 & 10.12 & 95.42 & 10.00 & 94.64 & 12.40 & 94.55 \\
Learnable (w/o DSSS) & 13.64 & 93.18 & 13.75 & 93.68 & 13.33 & 95.02 & 16.48 & 92.67 \\
\!\cellcolor{gray!9.0} Learnable (w DSSS) \!
& \!\cellcolor{gray!9.0} \textbf{10.00} \! & \!\cellcolor{gray!9.0} \textbf{95.70} \! &\!\cellcolor{gray!9.0} \textbf{6.50} \!& \!\cellcolor{gray!9.0} \textbf{97.98} \! &\!\cellcolor{gray!9.0} \textbf{7.91} \!& \!\cellcolor{gray!9.0} \textbf{96.83} \!& \!\cellcolor{gray!9.0} \textbf{8.14} \!& \!\cellcolor{gray!9.0} \textbf{96.49} \!\\
\bottomrule
\end{tabular}}
\end{center}
\vspace{-6mm}
\caption{Ablation studies on different selection strategies of style bases on four testing domains.}
\label{tab:ablation_selection}
\vspace{-1mm}
\end{table*}

\begin{table*}[h!]
\centering
\begin{center}
\vspace{-2mm}
\resizebox{1.0\textwidth}{!}{%
\begin{tabular}{c c c c  c c  c c  c c  c c }
\toprule
\multicolumn{4}{c}{\textbf{Loss}} &
\multicolumn{2}{c}{\textbf{I\&C\&M to O}} &
\multicolumn{2}{c}{\textbf{O\&C\&M to I}} &
\multicolumn{2}{c}{\textbf{O\&C\&I to M}} &
\multicolumn{2}{c}{\textbf{O\&M\&I to C}} \\
\!$\mathcal{L}_{\mathrm{cls}}$\! & \!$\mathcal{L}_{\mathrm{dep}}$\! &
\!$\mathcal{L}_{\mathrm{sty}}$\! & \!$\mathcal{L}_{\mathrm{con}}$\! 
& HTER(\%) &AUC(\%) &HTER(\%) &AUC(\%) &HTER(\%) &AUC(\%) &HTER(\%) &AUC(\%)\\
\midrule
\!\ding{51}\! & \!\textbf{--}\! & \!\textbf{--}\! & \!\textbf{--}\! 
& 15.96 & 90.77 & 16.40 & 91.65 & 16.16 & 92.44 & 18.53 & 89.77 \\
\!\ding{51}\! & \!\ding{51}\! & \!\textbf{--}\! & \!\textbf{--}\! 
& 13.64 & 93.18 & 13.75 & 93.68 & 13.33 & 95.02 & 16.48 & 92.67 \\
\!\ding{51}\! & \!\ding{51}\! & 
\!\ding{51}\! & \!\textbf{--}\! 
& 13.05 & 93.90 & 12.62 & 94.42 & 10.41 & 94.35 & 16.29 & 93.72 \\
\!\ding{51}\! & \!\ding{51}\! & 
 \!\textbf{--}\! & \!\ding{51}\! 
& 11.63 & 94.32 & 10.25 & 95.78 & 9.10 & 95.06 & 12.03 & 93.95 \\
\!\cellcolor{gray!9.0}\ding{51}\! & \!\cellcolor{gray!9.0}\ding{51}\!
& 
\!\cellcolor{gray!9.0}\ding{51}\! & \!\cellcolor{gray!9.0}\ding{51}\!
& \!\cellcolor{gray!9.0} \textbf{10.00} \! & \!\cellcolor{gray!9.0} \textbf{95.70} \! &\!\cellcolor{gray!9.0} \textbf{6.50} \!& \!\cellcolor{gray!9.0} \textbf{97.98} \! &\!\cellcolor{gray!9.0} \textbf{7.91} \!& \!\cellcolor{gray!9.0} \textbf{96.83} \!& \!\cellcolor{gray!9.0} \textbf{8.14} \!& \!\cellcolor{gray!9.0} \textbf{96.49} \!\\
\bottomrule
\end{tabular}
}
\end{center}
\vspace{-6mm}
\caption{Ablation studies on each loss (with TTSP) on four testing domains.}
\label{tab:ablation_loss}
\vspace{-4.7mm}
\end{table*}
\subsection{Comparisons to the State-of-the-art Methods}
\label{sec:4.2} 
\noindent \textbf{Comparison Results on Leave-One-Out Settings.} 
As shown in Table~\ref{tab:DG_SOTA_3to1}, Table~\ref{tab:test_time_DG}, we verify our proposed method in four standard leave-one-out settings. 
Note that in each experiment, all methods are compared using the same backbone to ensure fair comparisons. In all experiments, IADG~\cite{zhou2023instance} is implemented by using the same backbone that removes the DKG module.
From the tables, we draw the following observations. (1)  Our proposed TTDG method consistently outperforms the majority of state-of-the-art DG FAS methods~\cite{shao2019multi,jia2020single,shao2020regularized,liu2021adaptive,liu2021dual,du2022energy,zhou2022adaptive,liao2023domain} under five testing settings. 
This superiority can be attributed to the fact that most existing approaches tend to overlook the role that testing data plays in enhancing the generalizability of the FAS model beyond mere evaluation, resulting in sub-optimal performances. In contrast, we introduce test-time DG for FAS that leads to substantial improvements. (2) Existing test-time DG approaches~\cite{jiang2022domain,ijcai2022p240,huang2022adaptive} exhibit less-desired performances in these benchmark settings. The reasons lie in two aspects. Firstly, they tend to roughly define style bases, which is inapplicable to FAS tasks that typically have a mixture of domains. In contrast, we design $N$=64 learnable style bases to automatically capture style shifts in a fine-grained manner. Besides, their projections lack a clear direction for optimization, making projection less reliable for FAS. Conversely, we propose explicit optimization goals ($\mathcal{L}_{\mathrm{sty}}$\&$\mathcal{L}_{\mathrm{con}}$) to facilitate the projection process.

\noindent \textbf{Comparison Results on Limited Source Domains.} 
Following previous
works~\cite{shao2019multi,jia2020single,liu2021adaptive,liu2021dual,du2022energy}, we also evaluate our method on limited source
domains. 
Table~\ref{tab:DG_SOTA_2to1} shows that our method outperforms state-of-the-art approaches by a significant margin (5\% $\sim$ 6\% in HTER) when dealing with extremely limited source domains. This also demonstrates that our TTDG remains effective when applied to unseen domains, regardless of the number of source domains. 
In contrast to previous methods,
TTDG does not require any domain labels and is more flexible in realistic scenarios.

\subsection{Ablation Studies}
\label{sec:4.3}  
\noindent \textbf{Effects of Various Test-Time Style Shifting Schemes.} Table~\ref{tab:ablation_shifting} shows the effect of different test-time shifting schemes while preserving the DSSS unchanged. TF-Cal~\cite{ijcai2022p240} directly shifts the amplitude of the Fourier representation to the source prototype in a simple manner.
Similarly, TTSS~\cite{park2023test} shifts the style statistics of the test sample to the nearest source domain before making the prediction. However, they neglect the contribution of other similar source domains and achieve less desirable outcomes. 
Sty.-Pro~\cite{huang2023style} directly projects the unseen samples to bases via the Wasserstein distance~\cite{vallender1974calculation}, which cannot be directly applied to the hyper-spherical feature space we constructed and shows less-desired results in DG FAS (Table~\ref{tab:ablation_shifting}). In contrast, we introduce a cosine distance-based similarity (Eq. (\ref{eq:cosine_sim})) and perform the projection into the same space, which is more suitable for FAS task. 
Thus, our TTSP outperforms various test-time style shifting strategies by a large margin.

\noindent \textbf{Impacts of Different Style Bases Selection Strategies.} Table~\ref{tab:ablation_selection} illustrates the impact of various style bases selection strategies while keeping TTSP consistent. Random selection means randomly selecting $N$ style bases from all source domains. It shows inferior results in four domains since it cannot fully represent the style representation space. 
FPS~\cite{qi2017pointnet++} selects the bases according to the farthest point sampling strategy and achieves better results. However, both of them are time-expensive since they need to re-select the bases from all the source domains in each epoch. TTDG (w/o DSSS) means style bases are learned with task-related losses ($\mathcal{L}_{\mathrm{cls}}$ \& $\mathcal{L}_{\mathrm{dep}}$) only and such randomly learnable bases are diverse to some extent.
In contrast, our learnable bases under two proposed losses are more effective, further improving the performance by a large margin.

\vspace{-0.8mm}
\noindent \textbf{Contribution of Each Loss.} 
Table~\ref{tab:ablation_loss} demonstrates the contribution of each proposed loss with TTSP.
When we train style bases using $\mathcal{L}_{\mathrm{style}}$ but without $\mathcal{L}_{\mathrm{content}}$, we observe limited performance improvements compared to the baseline.
This is because the style-content features obtained become more diverse within the same class but lack content consistency. Conversely, merely using $\mathcal{L}_{\mathrm{content}}$ without $\mathcal{L}_{\mathrm{style}}$ achieves a certain performance boost since it encourages the style-content features to be more consistent with the content features but lacks style diversity.  Finally, only by jointly incorporating both losses will we achieve the best results. This shows that our TTDG needs
to be trained under the guidance of both loss functions ($\mathcal{L}_{\mathrm{style}}$ \& $\mathcal{L}_{\mathrm{content}}$).

\begin{figure}[t!]
\centering
\includegraphics[width=0.5\textwidth]{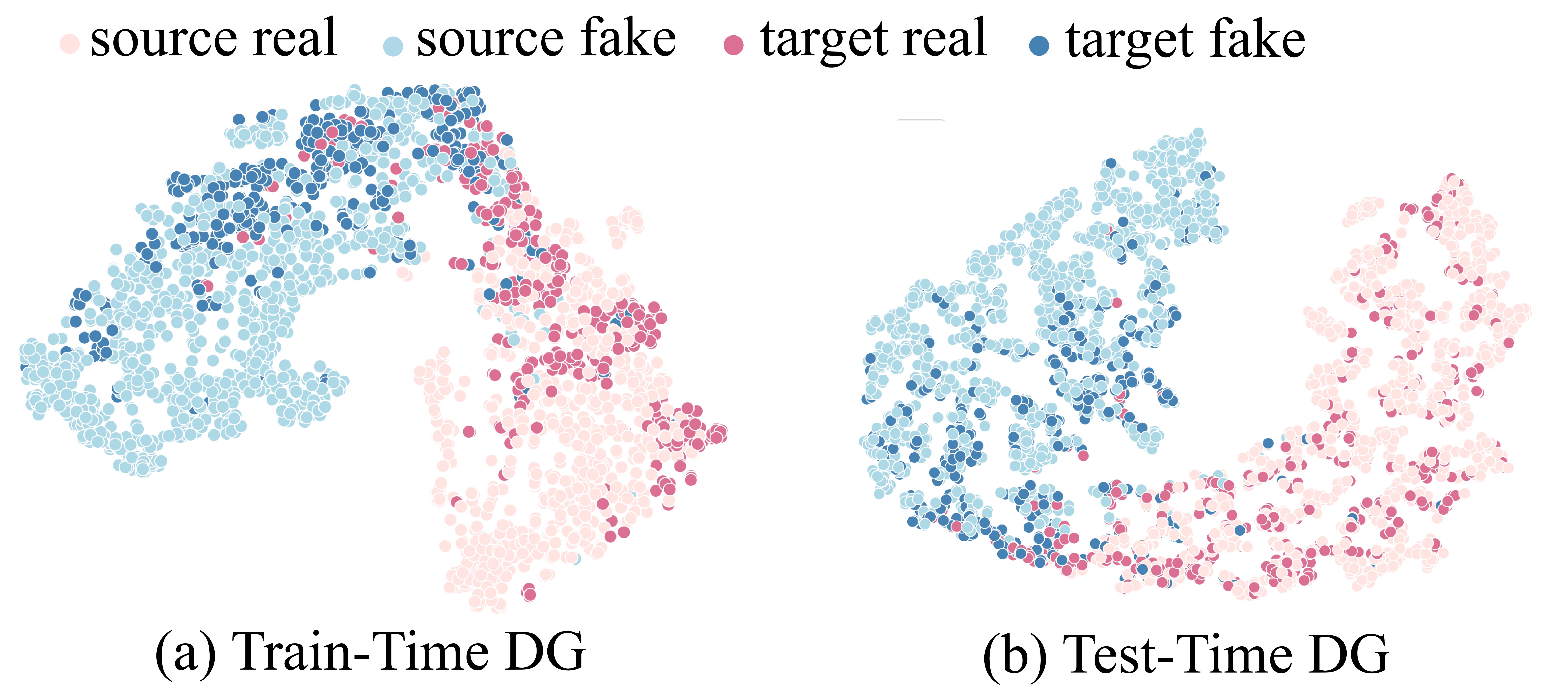}
\vspace{-5mm}
\caption{Comparison results of t-SNE~\cite{van2008visualizing} feature visualization for train-time DG and our test-time DG method. }
\label{fig:vis_tsne}
\end{figure}

\begin{figure}[t!]
\centering
\includegraphics[width=0.5\textwidth]{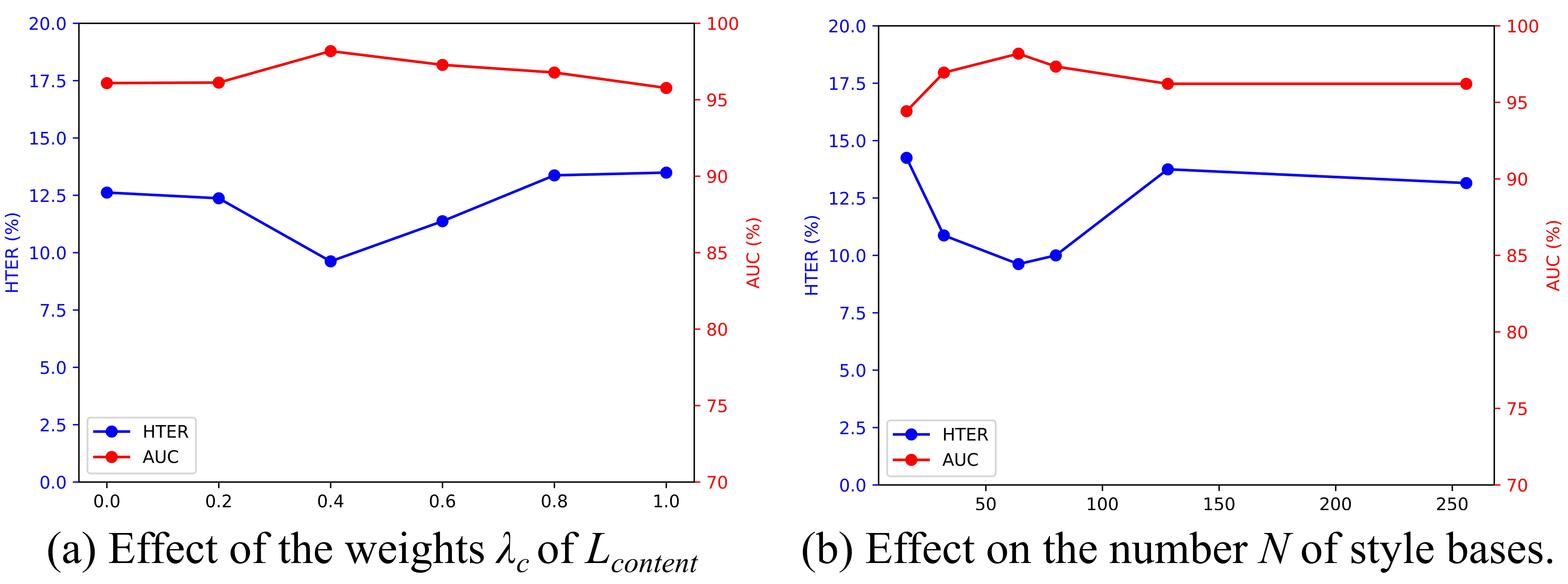}
\vspace{-6mm}
\caption{Hyper-parameter analyses on the O\&C\&M to I setting.}
\vspace{-4mm}
\label{fig:vis_params}
\end{figure}

\subsection{Visualization and Analysis}
\label{sec:4.4} 

\noindent \textbf{T-SNE Visualization of Feature Distributions.} To  reveal how testing data leads to the generalizability boost,
we employ the t-SNE~\cite{van2008visualizing} visualization tool on the I\&C\&M to O setting to analyze the effectiveness of our proposed method.

Figure~\ref{fig:vis_tsne} shows the feature distributions between train-time DG~\cite{zhou2022adaptive} and our test-time DG method. We have two observations: 
(1) In Figure~\ref{fig:vis_tsne} (a), testing samples near the decision boundary are almost misclassified, where the previous method is ineffective, while points away from the decision boundary are well-classified. (2) Although source samples in Figure~\ref{fig:vis_tsne} (a) are well-separated, many target samples are misclassified, while our method in Figure~\ref{fig:vis_tsne} (b) has much fewer misclassified samples, which verifies ours superiority. 
(3) In Figure~\ref{fig:vis_tsne} (b), our source and target domains exhibit better alignment, indicating better generalizability. This is because the target domain varies across samples, making the weights of the selected bases different in TTDG.

\begin{figure}[t!]
\centering
\includegraphics[width=0.5\textwidth]{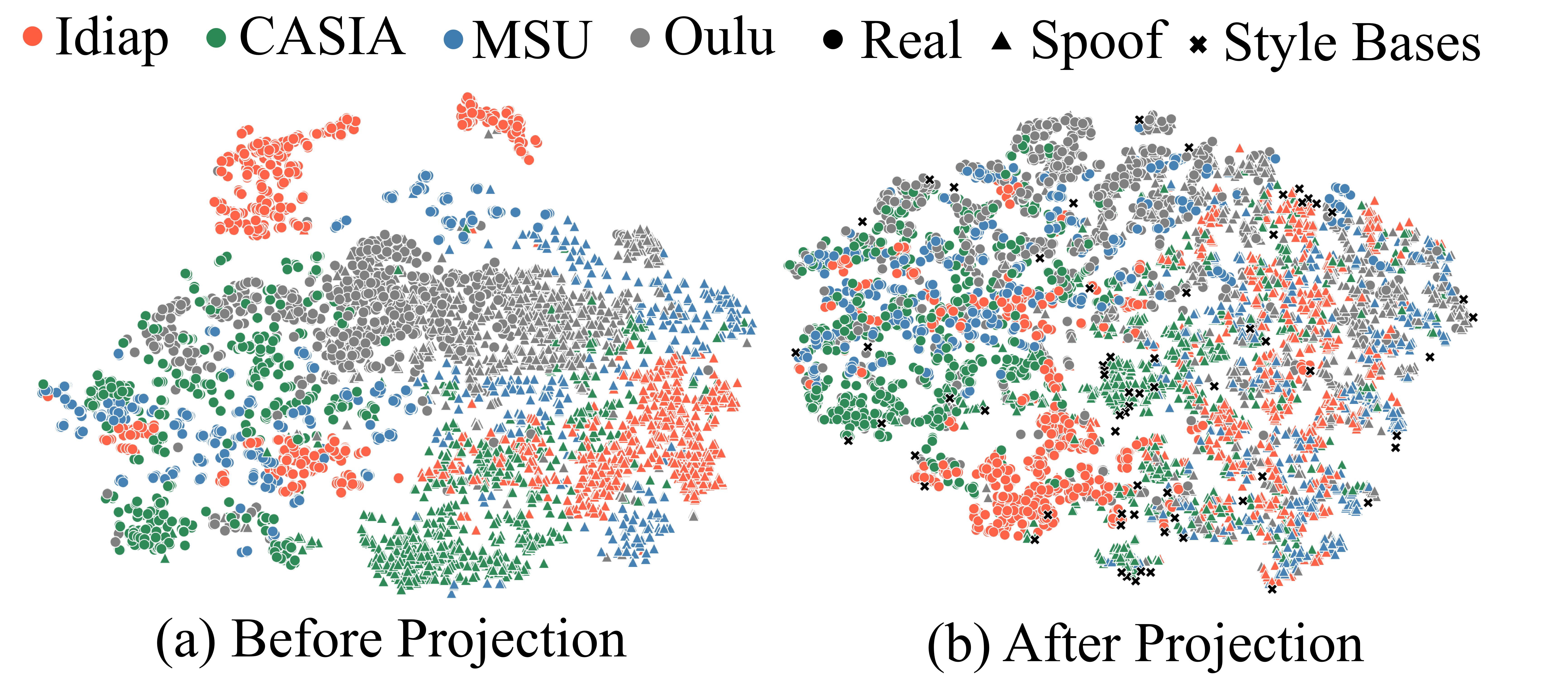}
\vspace{-5mm}
\caption{T-SNE~\cite{van2008visualizing} visualization of features for different 
domains before (a) and after test-time style projection (b). }
\vspace{-3mm}
\label{fig:vis_tsne_multi}
\end{figure}

Figure~\ref{fig:vis_tsne_multi}  illustrates the variations of style distributions between
different domains before and after style projection.
We have three observations as follows: (1)
Before style projection (Figure~\ref{fig:vis_tsne_multi} (a)), it is evident that the style distribution of distinct domains is separated. After style projection, the style distribution of the unseen domain is approximately situated within the style bases. (2) Furthermore, the unseen domain aligns more closely with the source domains (Figure~\ref{fig:vis_tsne_multi} (b)), 
demonstrating that TTSP successfully projects unseen styles into the seen space. (3) Finally, the learnable style bases are diverse enough to represent the whole space, and most of them lie in the outlier of the style representation space. When TTDG encounters an unseen sample (O), they often relate it to a previously perceived similar one (C), and thus some of the bases are near the domain C.

\noindent \textbf{Hyper-parameter Analysis.} During the optimization, it is essential to balance the weight between
different losses. 
We study the impact of $\lambda_c$ on TTDG-V.
As shown in Figure~\ref{fig:vis_params} (a), reducing $\lambda_c$ may not significantly facilitate the training process, while increasing it too much can result in the propagation of incorrect gradients throughout the network. Based on our empirical findings, we set $\lambda_c$ to 0.4 for all experiments. Next, we analyzed the impact on the number $N$ of style bases of TTDG-V in Figure~\ref{fig:vis_params} (b). A smaller value of $N$ is insufficient for representing the source style space, causing the model to become overly specific and resulting in poor generalization. Conversely, a larger value of $N$ introduces redundant bases, leading to less desirable outcomes. Thus, we set $N$ to a default value of 64 in experiments.

\section{Conclusion}
In this paper, we present a new perspective for DG FAS that leverages testing data to enhance the generalizability beyond mere evaluation. We propose a novel Test-Time Domain Generalization (TTDG) framework for FAS, which is the first work that studies test-time DG for FAS. Specifically, we introduce Test-Time Style Projection (TTSP) to project the styles of the unseen samples to the source domains via the aggregation of a set of style bases. In addition, we design Diverse Style Shifts Simulation (DSSS) to synthesize diverse distribution shifts via learnable style bases in a hyperspherical feature space, thereby promoting the test-time DG. 
Extensive experiments demonstrate the state-of-the-art performance and the effectiveness of our TTDG on widely used cross-domain FAS benchmarks.

\section{Acknowledgement}
The work is supported by Shanghai Municipal Science and Technology Major Project (2021SHZDZX0102), Shanghai Science and Technology Commission  (21511101200), National Natural Science Foundation of China (No. 72192821), YuCaiKe (No. 14105167-2023).
\newpage
\newpage
{
    \small
    \bibliographystyle{ieeenat_fullname}
    \bibliography{main}

\begin{thebibliography}{121}
\providecommand{\natexlab}[1]{#1}
\providecommand{\url}[1]{\texttt{#1}}
\expandafter\ifx\csname urlstyle\endcsname\relax
  \providecommand{\doi}[1]{doi: #1}\else
  \providecommand{\doi}{doi: \begingroup \urlstyle{rm}\Url}\fi

\bibitem[Albuquerque et~al.(2019)Albuquerque, Monteiro, Darvishi, Falk, and Mitliagkas]{albuquerque2019generalizing}
Isabela Albuquerque, Jo{\~a}o Monteiro, Mohammad Darvishi, Tiago~H Falk, and Ioannis Mitliagkas.
\newblock Generalizing to unseen domains via distribution matching.
\newblock \emph{arXiv preprint arXiv:1911.00804}, 2019.

\bibitem[Atoum et~al.(2017)Atoum, Liu, Jourabloo, and Liu]{atoum2017face}
Yousef Atoum, Yaojie Liu, Amin Jourabloo, and Xiaoming Liu.
\newblock Face anti-spoofing using patch and depth-based cnns.
\newblock In \emph{IEEE International Joint Conference on Biometrics (IJCB)}, pages 319--328, 2017.

\bibitem[Ben-David et~al.(2010)Ben-David, Blitzer, Crammer, Kulesza, Pereira, and Vaughan]{ben2010theory}
Shai Ben-David, John Blitzer, Koby Crammer, Alex Kulesza, Fernando Pereira, and Jennifer~Wortman Vaughan.
\newblock A theory of learning from different domains.
\newblock \emph{Machine learning (ML)}, 79:\penalty0 151--175, 2010.

\bibitem[Boulkenafet et~al.(2015)Boulkenafet, Komulainen, Hadid, et~al.]{boulkenafet2015face}
Zinelabidine Boulkenafet, Jukka Komulainen, Abdenour Hadid, et~al.
\newblock Face anti-spoofing based on color texture analysis.
\newblock In \emph{IEEE International Conference on Image Processing (ICIP)}, pages 2636--2640, 2015.

\bibitem[Boulkenafet et~al.(2017{\natexlab{a}})Boulkenafet, Komulainen, Hadid, et~al.]{2017Face}
Zinelabidine Boulkenafet, Jukka Komulainen, Abdenour Hadid, et~al.
\newblock Face spoofing detection using colour texture analysis.
\newblock \emph{IEEE Transactions on Information Forensics and Security (TIFS)}, 11\penalty0 (8):\penalty0 1818--1830, 2017{\natexlab{a}}.

\bibitem[Boulkenafet et~al.(2017{\natexlab{b}})Boulkenafet, Komulainen, Li, Feng, and Hadid]{2017OULU}
Zinelabinde Boulkenafet, Jukka Komulainen, Lei Li, Xiaoyi Feng, and Abdenour Hadid.
\newblock Oulu-npu: A mobile face presentation attack database with real-world variations.
\newblock In \emph{12th IEEE International Conference on Automatic Face \& Gesture Recognition (FG)}, pages 612--618, 2017{\natexlab{b}}.

\bibitem[Cai et~al.(2022)Cai, Li, Wan, Li, Hu, and Kot]{cai2022learning}
Rizhao Cai, Zhi Li, Renjie Wan, Haoliang Li, Yongjian Hu, and Alex~C Kot.
\newblock Learning meta pattern for face anti-spoofing.
\newblock \emph{IEEE Transactions on Information Forensics and Security (TIFS)}, 17:\penalty0 1201--1213, 2022.

\bibitem[Chen et~al.(2022)Chen, Wang, Darrell, and Ebrahimi]{chen2022contrastive}
Dian Chen, Dequan Wang, Trevor Darrell, and Sayna Ebrahimi.
\newblock Contrastive test-time adaptation.
\newblock In \emph{Proceedings of the IEEE/CVF Conference on Computer Vision and Pattern Recognition (CVPR)}, pages 295--305, 2022.

\bibitem[Chen et~al.(2021)Chen, Yao, Sheng, Ding, Tai, Li, Huang, and Jin]{chen2021generalizable}
Zhihong Chen, Taiping Yao, Kekai Sheng, Shouhong Ding, Ying Tai, Jilin Li, Feiyue Huang, and Xinyu Jin.
\newblock Generalizable representation learning for mixture domain face anti-spoofing.
\newblock In \emph{Proceedings of the AAAI Conference on Artificial Intelligence (AAAI)}, pages 1132--1139, 2021.

\bibitem[Chingovska et~al.(2012)Chingovska, Anjos, Marcel, et~al.]{2012Replay}
Ivana Chingovska, Andr{\'e} Anjos, S{\'e}bastien Marcel, et~al.
\newblock On the effectiveness of local binary patterns in face anti-spoofing.
\newblock In \emph{International Conference of Biometrics Special Interest Group (BIOSIG)}, pages 1--7, 2012.

\bibitem[Deng et~al.(2019)Deng, Guo, Xue, and Zafeiriou]{deng2019arcface}
Jiankang Deng, Jia Guo, Niannan Xue, and Stefanos Zafeiriou.
\newblock Arcface: Additive angular margin loss for deep face recognition.
\newblock In \emph{Proceedings of the IEEE/CVF Conference on Computer Vision and Pattern Recognition (CVPR)}, pages 4690--4699, 2019.

\bibitem[Dosovitskiy et~al.(2020)Dosovitskiy, Beyer, Kolesnikov, Weissenborn, Zhai, Unterthiner, Dehghani, Minderer, Heigold, Gelly, et~al.]{dosovitskiy2020image}
Alexey Dosovitskiy, Lucas Beyer, Alexander Kolesnikov, Dirk Weissenborn, Xiaohua Zhai, Thomas Unterthiner, Mostafa Dehghani, Matthias Minderer, Georg Heigold, Sylvain Gelly, et~al.
\newblock An image is worth 16x16 words: Transformers for image recognition at scale.
\newblock In \emph{International Conference on Learning Representations (ICLR)}, 2020.

\bibitem[Du et~al.(2022)Du, Li, Zuo, Zhu, and Lu]{du2022energy}
Zhekai Du, Jingjing Li, Lin Zuo, Lei Zhu, and Ke Lu.
\newblock Energy-based domain generalization for face anti-spoofing.
\newblock In \emph{Proceedings of the 30th ACM International Conference on Multimedia (ACM MM)}, pages 1749--1757, 2022.

\bibitem[Feng et~al.(2016)Feng, Po, Li, Xu, Yuan, Cheung, and Cheung]{DeepBinary00}
Litong Feng, Lai-Man Po, Yuming Li, Xuyuan Xu, Fang Yuan, Terence Chun-Ho Cheung, and Kwok-Wai Cheung.
\newblock Integration of image quality and motion cues for face anti-spoofing: A neural network approach.
\newblock \emph{Journal of Visual Communication and Image Representation (JVCIR)}, 38:\penalty0 451--460, 2016.

\bibitem[Feng et~al.(2018)Feng, Wu, Shao, Wang, and Zhou]{2018Joint3D}
Yao Feng, Fan Wu, Xiaohu Shao, Yanfeng Wang, and Xi Zhou.
\newblock Joint 3d face reconstruction and dense alignment with position map regression network.
\newblock In \emph{European Conference on Computer Vision (ECCV)}, pages 534--551, 2018.

\bibitem[Feng et~al.(2022)Feng, Zhou, Gu, Tan, Cheng, Lu, Shi, and Ma]{feng2020dmt}
Zhengyang Feng, Qianyu Zhou, Qiqi Gu, Xin Tan, Guangliang Cheng, Xuequan Lu, Jianping Shi, and Lizhuang Ma.
\newblock Dmt: Dynamic mutual training for semi-supervised learning.
\newblock \emph{Pattern Recognition (PR)}, page 108777, 2022.

\bibitem[Freitas~Pereira et~al.(2014)Freitas~Pereira, Komulainen, Anjos, De~Martino, Hadid, Pietikäinen, and Marcel]{2014dynamic}
Tiago Freitas~Pereira, Jukka Komulainen, André Anjos, José De~Martino, Abdenour Hadid, Matti Pietikäinen, and Sébastien Marcel.
\newblock Face liveness detection using dynamic texture.
\newblock \emph{Eurasip Journal on Image and Video Processing (JIVP)}, 2014\penalty0 (1):\penalty0 1--15, 2014.

\bibitem[Freitas~Pereira et~al.(2012)Freitas~Pereira, Anjos, Martino, and Marcel]{LBP01}
Tiago~de Freitas~Pereira, Andr{\'e} Anjos, Jos{\'e} Mario~De Martino, and S{\'e}bastien Marcel.
\newblock Lbp- top based countermeasure against face spoofing attacks.
\newblock In \emph{Asian Conference on Computer Vision (ACCV)}, pages 121--132, 2012.

\bibitem[Ganin and Lempitsky(2015)]{ganin2015unsupervised}
Yaroslav Ganin and Victor Lempitsky.
\newblock Unsupervised domain adaptation by backpropagation.
\newblock In \emph{International Conference on Machine Learning (ICML)}, pages 1180--1189, 2015.

\bibitem[George and Marcel(2021)]{george2021effectiveness}
Anjith George and S{\'e}bastien Marcel.
\newblock On the effectiveness of vision transformers for zero-shot face anti-spoofing.
\newblock In \emph{IEEE International Joint Conference on Biometrics (IJCB)}, pages 1--8, 2021.

\bibitem[Gu et~al.(2021)Gu, Zhou, Xu, Feng, Cheng, Lu, Shi, and Ma]{PIT}
Qiqi Gu, Qianyu Zhou, Minghao Xu, Zhengyang Feng, Guangliang Cheng, Xuequan Lu, Jianping Shi, and Lizhuang Ma.
\newblock Pit: Position-invariant transform for cross-fov domain adaptation.
\newblock In \emph{Proceedings of the IEEE/CVF International Conference on Computer Vision (ICCV)}, pages 8761--8770, 2021.

\bibitem[Guo et~al.(2021)Guo, Zhou, Zhou, Gu, Tang, Feng, and Ma]{guo2021label}
Shaohua Guo, Qianyu Zhou, Ye Zhou, Qiqi Gu, Junshu Tang, Zhengyang Feng, and Lizhuang Ma.
\newblock Label-free regional consistency for image-to-image translation.
\newblock In \emph{IEEE International Conference on Multimedia and Expo (ICME)}, pages 1--6, 2021.

\bibitem[Hong et~al.(2023)Hong, Lin, Liu, Yeh, and Chen]{hong2023domain}
Zong-Wei Hong, Yu-Chen Lin, Hsuan-Tung Liu, Yi-Ren Yeh, and Chu-Song Chen.
\newblock Domain-generalized face anti-spoofing with unknown attacks.
\newblock In \emph{IEEE International Conference on Image Processing (ICIP)}, pages 820--824, 2023.

\bibitem[Hu et~al.(2022)Hu, Cao, Zhang, Yao, Ding, and Ma]{hu2022structure}
Chengyang Hu, Junyi Cao, Ke-Yue Zhang, Taiping Yao, Shouhong Ding, and Lizhuang Ma.
\newblock Structure destruction and content combination for generalizable anti-spoofing.
\newblock \emph{IEEE Transactions on Biometrics, Behavior, and Identity Science (TBIOM)}, 4\penalty0 (4):\penalty0 508--521, 2022.

\bibitem[Hu et~al.(2024{\natexlab{a}})Hu, Zhang, Yao, Ding, and Ma]{hu2024rethinking}
Chengyang Hu, Ke-Yue Zhang, Taiping Yao, Shouhong Ding, and Lizhuang Ma.
\newblock Rethinking generalizable face anti-spoofing via hierarchical prototype-guided distribution refinement in hyperbolic space.
\newblock In \emph{Proceedings of the IEEE/CVF Conference on Computer Vision and Pattern Recognition (CVPR)}, 2024{\natexlab{a}}.

\bibitem[Hu et~al.(2024{\natexlab{b}})Hu, Zhang, Yao, Liu, Ding, Tan, and Ma]{hu2024domain}
Chengyang Hu, Ke-Yue Zhang, Taiping Yao, Shice Liu, Shouhong Ding, Xin Tan, and Lizhuang Ma.
\newblock Domain-hallucinated updating for multi-domain face anti-spoofing.
\newblock In \emph{Proceedings of the AAAI Conference on Artificial Intelligence (AAAI)}, pages 2193--2201, 2024{\natexlab{b}}.

\bibitem[Huang et~al.(2022)Huang, Sun, Liu, Chu, Xiao, Yuan, Adam, and Yang]{huang2022adaptive}
Hsin-Ping Huang, Deqing Sun, Yaojie Liu, Wen-Sheng Chu, Taihong Xiao, Jinwei Yuan, Hartwig Adam, and Ming-Hsuan Yang.
\newblock Adaptive transformers for robust few-shot cross-domain face anti-spoofing.
\newblock In \emph{European Conference on Computer Vision (ECCV)}, pages 37--54, 2022.

\bibitem[Huang et~al.(2023{\natexlab{a}})Huang, Chiang, Chong, Chen, Ni, and Hsu]{huang2023ldcformer}
Pei-Kai Huang, Cheng-Hsuan Chiang, Jun-Xiong Chong, Tzu-Hsien Chen, Hui-Yu Ni, and Chiou-Ting Hsu.
\newblock Ldcformer: Incorporating learnable descriptive convolution to vision transformer for face anti-spoofing.
\newblock In \emph{IEEE International Conference on Image Processing (ICIP)}, pages 121--125, 2023{\natexlab{a}}.

\bibitem[Huang et~al.(2023{\natexlab{b}})Huang, Chen, Li, Li, Li, Song, Yan, and Xiong]{huang2023style}
Wei Huang, Chang Chen, Yong Li, Jiacheng Li, Cheng Li, Fenglong Song, Youliang Yan, and Zhiwei Xiong.
\newblock Style projected clustering for domain generalized semantic segmentation.
\newblock In \emph{Proceedings of the IEEE/CVF Conference on Computer Vision and Pattern Recognition (CVPR)}, pages 3061--3071, 2023{\natexlab{b}}.

\bibitem[Jia et~al.(2020)Jia, Zhang, Shan, and Chen]{jia2020single}
Yunpei Jia, Jie Zhang, Shiguang Shan, and Xilin Chen.
\newblock Single-side domain generalization for face anti-spoofing.
\newblock In \emph{Proceedings of the IEEE/CVF Conference on Computer Vision and Pattern Recognition (CVPR)}, pages 8484--8493, 2020.

\bibitem[Jia et~al.(2021{\natexlab{a}})Jia, Zhang, and Shan]{jia2021dual}
Yunpei Jia, Jie Zhang, and Shiguang Shan.
\newblock Dual-branch meta-learning network with distribution alignment for face anti-spoofing.
\newblock \emph{IEEE Transactions on Information Forensics and Security (TIFS)}, 17:\penalty0 138--151, 2021{\natexlab{a}}.

\bibitem[Jia et~al.(2021{\natexlab{b}})Jia, Zhang, Shan, and Chen]{jia2021unified}
Yunpei Jia, Jie Zhang, Shiguang Shan, and Xilin Chen.
\newblock Unified unsupervised and semi-supervised domain adaptation network for cross-scenario face anti-spoofing.
\newblock \emph{Pattern Recognition (PR)}, 115:\penalty0 107888, 2021{\natexlab{b}}.

\bibitem[Jiang et~al.(2022)Jiang, Wang, Zhang, Xu, Zhang, Chen, and Tian]{jiang2022domain}
Yuxuan Jiang, Yanfeng Wang, Ruipeng Zhang, Qinwei Xu, Ya Zhang, Xin Chen, and Qi Tian.
\newblock Domain-conditioned normalization for test-time domain generalization.
\newblock In \emph{European Conference on Computer Vision (ECCV)}, pages 291--307, 2022.

\bibitem[Jourabloo et~al.(2018)Jourabloo, Liu, and Liu]{jourabloo2018face}
Amin Jourabloo, Yaojie Liu, and Xiaoming Liu.
\newblock Face de-spoofing: Anti-spoofing via noise modeling.
\newblock In \emph{European Conference on Computer Vision (ECCV)}, pages 290--306, 2018.

\bibitem[Kemelmacher-Shlizerman et~al.(2016)Kemelmacher-Shlizerman, Seitz, Miller, and Brossard]{kemelmacher2016megaface}
Ira Kemelmacher-Shlizerman, Steven~M Seitz, Daniel Miller, and Evan Brossard.
\newblock The megaface benchmark: 1 million faces for recognition at scale.
\newblock In \emph{Proceedings of the IEEE Conference on Computer Vision and Pattern Recognition (CVPR)}, pages 4873--4882, 2016.

\bibitem[Kim et~al.(2019)Kim, Kim, Kim, and Kim]{kim2019basn}
Taewook Kim, YongHyun Kim, Inhan Kim, and Daijin Kim.
\newblock Basn: Enriching feature representation using bipartite auxiliary supervisions for face anti-spoofing.
\newblock In \emph{Proceedings of the IEEE/CVF International Conference on Computer Vision Workshops (ICCVW)}, pages 494--503, 2019.

\bibitem[Kim et~al.(2022)Kim, Nam, Min, and Lee]{kim2022style}
Young-Eun Kim, Woo-Jeoung Nam, Kyungseo Min, and Seong-Whan Lee.
\newblock Style-guided domain adaptation for face presentation attack detection.
\newblock \emph{arXiv preprint arXiv:2203.14565}, 2022.

\bibitem[Kim et~al.(2023)Kim, Nam, Min, and Lee]{kim2023style}
Young-Eun Kim, Woo-Jeoung Nam, Kyungseo Min, and Seong-Whan Lee.
\newblock Style selective normalization with meta learning for test-time adaptive face anti-spoofing.
\newblock \emph{Expert Systems with Applications (ESWA)}, 214:\penalty0 119106, 2023.

\bibitem[Komulainen et~al.(2013)Komulainen, Hadid, Pietik{\"a}inen, et~al.]{2014Context}
Jukka Komulainen, Abdenour Hadid, Matti Pietik{\"a}inen, et~al.
\newblock Context based face anti-spoofing.
\newblock In \emph{IEEE Sixth International Conference on Biometrics: Theory, Applications and Systems (BTAS)}, pages 1--8, 2013.

\bibitem[Kwak et~al.(2023)Kwak, Jung, Yoo, Shin, and Kim]{kwak2023liveness}
Youngjun Kwak, Minyoung Jung, Hunjae Yoo, JinHo Shin, and Changick Kim.
\newblock Liveness score-based regression neural networks for face anti-spoofing.
\newblock In \emph{IEEE International Conference on Acoustics, Speech and Signal Processing (ICASSP)}, pages 1--5, 2023.

\bibitem[Le and Woo(2024)]{le2024gradient}
Binh~M Le and Simon~S Woo.
\newblock Gradient alignment for cross-domain face anti-spoofing.
\newblock In \emph{Proceedings of the IEEE/CVF Conference on Computer Vision and Pattern Recognition (CVPR)}, 2024.

\bibitem[Li et~al.(2018{\natexlab{a}})Li, Yang, Song, and Hospedales]{li2018learning}
Da Li, Yongxin Yang, Yi-Zhe Song, and Timothy Hospedales.
\newblock Learning to generalize: Meta-learning for domain generalization.
\newblock In \emph{Proceedings of the AAAI conference on artificial intelligence (AAAI)}, pages 3490--3497, 2018{\natexlab{a}}.

\bibitem[Li et~al.(2018{\natexlab{b}})Li, Li, Cao, Wang, Huang, and Kot]{2018Unsupervised}
Haoliang Li, Wen Li, Hong Cao, Shiqi Wang, Feiyue Huang, and Alex~C. Kot.
\newblock Unsupervised domain adaptation for face anti-spoofing.
\newblock \emph{IEEE Transactions on Information Forensics and Security (TIFS)}, 13\penalty0 (7):\penalty0 1794--1809, 2018{\natexlab{b}}.

\bibitem[Li et~al.(2018{\natexlab{c}})Li, Pan, Wang, and Kot]{li2018domain}
Haoliang Li, Sinno~Jialin Pan, Shiqi Wang, and Alex~C Kot.
\newblock Domain generalization with adversarial feature learning.
\newblock In \emph{Proceedings of the IEEE Conference on Computer Vision and Pattern Recognition (CVPR)}, pages 5400--5409, 2018{\natexlab{c}}.

\bibitem[Li et~al.(2016)Li, Feng, Boulkenafet, Xia, Li, and Hadid]{DeepBinary01}
Lei Li, Xiaoyi Feng, Zinelabidine Boulkenafet, Zhaoqiang Xia, Mingming Li, and Abdenour Hadid.
\newblock An original face anti-spoofing approach using partial convolutional neural network.
\newblock In \emph{International Conference on Image Processing Theory, Tools and Applications (IPTA)}, pages 1--6, 2016.

\bibitem[Li et~al.(2021)Li, Xu, Xu, Shen, Li, and Hooi]{li2021spherical}
Shen Li, Jianqing Xu, Xiaqing Xu, Pengcheng Shen, Shaoxin Li, and Bryan Hooi.
\newblock Spherical confidence learning for face recognition.
\newblock In \emph{Proceedings of the IEEE/CVF Conference on Computer Vision and Pattern Recognition (CVPR)}, pages 15629--15637, 2021.

\bibitem[Liao et~al.(2023)Liao, Chen, Liu, Yeh, Hu, and Chen]{liao2023domain}
Chen-Hao Liao, Wen-Cheng Chen, Hsuan-Tung Liu, Yi-Ren Yeh, Min-Chun Hu, and Chu-Song Chen.
\newblock Domain invariant vision transformer learning for face anti-spoofing.
\newblock In \emph{Proceedings of the IEEE/CVF Winter Conference on Applications of Computer Vision (WACV)}, pages 6098--6107, 2023.

\bibitem[Lin et~al.(2019)Lin, Li, Yu, and Zhao]{lin2019face}
Bofan Lin, Xiaobai Li, Zitong Yu, and Guoying Zhao.
\newblock Face liveness detection by rppg features and contextual patch-based cnn.
\newblock In \emph{Proceedings of the 3rd International Conference on Biometric Engineering and Applications (ICBEA)}, pages 61--68, 2019.

\bibitem[Lin et~al.(2024)Lin, Wang, Cai, Liu, Fu, Yu, Tang, and Kot]{lin2024suppress}
Xun Lin, Shuai Wang, Rizhao Cai, Yizhong Liu, Ying Fu, Zitong Yu, Wenzhong Tang, and Alex Kot.
\newblock Suppress and rebalance: Towards generalized multi-modal face anti-spoofing.
\newblock In \emph{Proceedings of the IEEE/CVF Conference on Computer Vision and Pattern Recognition (CVPR)}, 2024.

\bibitem[Liu and Liang(2022)]{liu2023ma}
Ajian Liu and Yanyan Liang.
\newblock Ma-vit: Modality-agnostic vision transformers for face anti-spoofing.
\newblock In \emph{Proceedings of the Thirty-First International Joint Conference on Artificial Intelligence, (IJCAI)}, pages 1180--1186, 2022.

\bibitem[Liu et~al.(2021{\natexlab{a}})Liu, Tan, Wan, Liang, Lei, Guo, and Li]{liu2021face}
Ajian Liu, Zichang Tan, Jun Wan, Yanyan Liang, Zhen Lei, Guodong Guo, and Stan~Z Li.
\newblock Face anti-spoofing via adversarial cross-modality translation.
\newblock \emph{IEEE Transactions on Information Forensics and Security (TIFS)}, 16:\penalty0 2759--2772, 2021{\natexlab{a}}.

\bibitem[Liu et~al.(2023{\natexlab{a}})Liu, Tan, Yu, Zhao, Wan, Lei, Zhang, Li, and Guo]{liu2023fm}
Ajian Liu, Zichang Tan, Zitong Yu, Chenxu Zhao, Jun Wan, Yanyan Liang~Zhen Lei, Du Zhang, Stan~Z Li, and Guodong Guo.
\newblock Fm-vit: Flexible modal vision transformers for face anti-spoofing.
\newblock \emph{IEEE Transactions on Information Forensics and Security (TIFS)}, 18:\penalty0 4775--4786, 2023{\natexlab{a}}.

\bibitem[Liu et~al.(2024{\natexlab{a}})Liu, Gong, Zhou, Lu, Yi, Xie, and Ma]{liu2024cloud}
Fengqi Liu, Jingyu Gong, Qianyu Zhou, Xuequan Lu, Ran Yi, Yuan Xie, and Lizhuang Ma.
\newblock Cloudmix: Dual mixup consistency for unpaired point cloud completion.
\newblock \emph{IEEE Transactions on Visualization and Computer Graphics (TVCG)}, 2024{\natexlab{a}}.

\bibitem[Liu et~al.(2021{\natexlab{b}})Liu, Zhang, Yao, Bi, Ding, Li, Huang, and Ma]{liu2021adaptive}
Shubao Liu, Ke-Yue Zhang, Taiping Yao, Mingwei Bi, Shouhong Ding, Jilin Li, Feiyue Huang, and Lizhuang Ma.
\newblock Adaptive normalized representation learning for generalizable face anti-spoofing.
\newblock In \emph{Proceedings of the 29th ACM International Conference on Multimedia (ACM MM)}, pages 1469--1477, 2021{\natexlab{b}}.

\bibitem[Liu et~al.(2021{\natexlab{c}})Liu, Zhang, Yao, Sheng, Ding, Tai, Li, Xie, and Ma]{liu2021dual}
Shubao Liu, Ke-Yue Zhang, Taiping Yao, Kekai Sheng, Shouhong Ding, Ying Tai, Jilin Li, Yuan Xie, and Lizhuang Ma.
\newblock Dual reweighting domain generalization for face presentation attack detection.
\newblock In \emph{Proceedings of the Thirtieth International Joint Conference on Artificial Intelligence (IJCAI)}, pages 867--873, 2021{\natexlab{c}}.

\bibitem[Liu et~al.(2022{\natexlab{a}})Liu, Lu, Xu, Yang, Ding, and Ma]{liu2022feature}
Shice Liu, Shitao Lu, Hongyi Xu, Jing Yang, Shouhong Ding, and Lizhuang Ma.
\newblock Feature generation and hypothesis verification for reliable face anti-spoofing.
\newblock In \emph{Proceedings of the AAAI Conference on Artificial Intelligence (AAAI)}, pages 1782--1791, 2022{\natexlab{a}}.

\bibitem[Liu and Liu(2022)]{liu2022spoof}
Yaojie Liu and Xiaoming Liu.
\newblock Spoof trace disentanglement for generic face anti-spoofing.
\newblock \emph{IEEE Transactions on Pattern Analysis and Machine Intelligence (TPAMI)}, 45\penalty0 (3):\penalty0 3813--3830, 2022.

\bibitem[Liu et~al.(2018)Liu, Jourabloo, Liu, et~al.]{2018Learning}
Yaojie Liu, Amin Jourabloo, Xiaoming Liu, et~al.
\newblock Learning deep models for face anti-spoofing: Binary or auxiliary supervision.
\newblock In \emph{Proceedings of the IEEE Conference on Computer Vision and Pattern Recognition (CVPR)}, pages 389--398, 2018.

\bibitem[Liu et~al.(2019)Liu, Stehouwer, Jourabloo, and Liu]{liu2019deep}
Yaojie Liu, Joel Stehouwer, Amin Jourabloo, and Xiaoming Liu.
\newblock Deep tree learning for zero-shot face anti-spoofing.
\newblock In \emph{Proceedings of the IEEE/CVF Conference on Computer Vision and Pattern Recognition (CVPR)}, pages 4680--4689, 2019.

\bibitem[Liu et~al.(2020)Liu, Stehouwer, Liu, et~al.]{STCN}
Yaojie Liu, Joel Stehouwer, Xiaoming Liu, et~al.
\newblock On disentangling spoof trace for generic face anti-spoofing.
\newblock In \emph{European Conference on Computer Vision (ECCV)}, pages 406--422, 2020.

\bibitem[Liu et~al.(2022{\natexlab{b}})Liu, Chen, Dai, Gou, Huang, and Xiong]{liu2022source}
Yuchen Liu, Yabo Chen, Wenrui Dai, Mengran Gou, Chun-Ting Huang, and Hongkai Xiong.
\newblock Source-free domain adaptation with contrastive domain alignment and self-supervised exploration for face anti-spoofing.
\newblock In \emph{European Conference on Computer Vision (ECCV)}, pages 511--528, 2022{\natexlab{b}}.

\bibitem[Liu et~al.(2022{\natexlab{c}})Liu, Chen, Dai, Li, Zou, and Xiong]{liu2022causal}
Yuchen Liu, Yabo Chen, Wenrui Dai, Chenglin Li, Junni Zou, and Hongkai Xiong.
\newblock Causal intervention for generalizable face anti-spoofing.
\newblock In \emph{IEEE International Conference on Multimedia and Expo (ICME)}, pages 01--06, 2022{\natexlab{c}}.

\bibitem[Liu et~al.(2023{\natexlab{b}})Liu, Chen, Gou, Huang, Wang, Dai, and Xiong]{liu2023towards}
Yuchen Liu, Yabo Chen, Mengran Gou, Chun-Ting Huang, Yaoming Wang, Wenrui Dai, and Hongkai Xiong.
\newblock Towards unsupervised domain generalization for face anti-spoofing.
\newblock In \emph{Proceedings of the IEEE/CVF International Conference on Computer Vision (ICCV)}, pages 20654--20664, 2023{\natexlab{b}}.

\bibitem[Liu et~al.(2024{\natexlab{b}})Liu, Chen, Dai, Gou, Huang, and Xiong]{liu2024source}
Yuchen Liu, Yabo Chen, Wenrui Dai, Mengran Gou, Chun-Ting Huang, and Hongkai Xiong.
\newblock Source-free domain adaptation with domain generalized pretraining for face anti-spoofing.
\newblock \emph{IEEE Transactions on Pattern Analysis and Machine Intelligence (TPAMI)}, 2024{\natexlab{b}}.

\bibitem[Long et~al.(2023{\natexlab{a}})Long, Zhou, Ying, Ma, and Luo]{long2023diverse}
Shaocong Long, Qianyu Zhou, Chenhao Ying, Lizhuang Ma, and Yuan Luo.
\newblock Diverse target and contribution scheduling for domain generalization.
\newblock \emph{arXiv preprint arXiv:2309.16460}, 2023{\natexlab{a}}.

\bibitem[Long et~al.(2023{\natexlab{b}})Long, Zhou, Ying, Ma, and Luo]{long2023rethink}
Shaocong Long, Qianyu Zhou, Chenhao Ying, Lizhuang Ma, and Yuan Luo.
\newblock Rethinking domain generalization: Discriminability and generalizability.
\newblock \emph{arXiv preprint arXiv:2309.16483}, 2023{\natexlab{b}}.

\bibitem[M{\"a}{\"a}tt{\"a} et~al.(2011)M{\"a}{\"a}tt{\"a}, Hadid, and Pietik{\"a}inen]{maatta2011face}
Jukka M{\"a}{\"a}tt{\"a}, Abdenour Hadid, and Matti Pietik{\"a}inen.
\newblock Face spoofing detection from single images using micro-texture analysis.
\newblock In \emph{International Joint Conference on Biometrics (IJCB)}, pages 1--7, 2011.

\bibitem[Mehta and Rastegari(2021)]{mehta2021mobilevit}
Sachin Mehta and Mohammad Rastegari.
\newblock Mobilevit: light-weight, general-purpose, and mobile-friendly vision transformer.
\newblock \emph{arXiv preprint arXiv:2110.02178}, 2021.

\bibitem[Niu et~al.(2022)Niu, Wu, Zhang, Chen, Zheng, Zhao, and Tan]{niu2022efficient}
Shuaicheng Niu, Jiaxiang Wu, Yifan Zhang, Yaofo Chen, Shijian Zheng, Peilin Zhao, and Mingkui Tan.
\newblock Efficient test-time model adaptation without forgetting.
\newblock In \emph{International Conference on Machine Learning (ICML)}, pages 16888--16905, 2022.

\bibitem[Panwar et~al.(2021)Panwar, Singh, Saha, Paudel, and Van~Gool]{panwar2021unsupervised}
Ankush Panwar, Pratyush Singh, Suman Saha, Danda~Pani Paudel, and Luc Van~Gool.
\newblock Unsupervised compound domain adaptation for face anti-spoofing.
\newblock In \emph{16th IEEE International Conference on Automatic Face and Gesture Recognition (FG)}, pages 1--8, 2021.

\bibitem[Park et~al.(2023)Park, Han, Kim, and Moon]{park2023test}
Jungwuk Park, Dong-Jun Han, Soyeong Kim, and Jaekyun Moon.
\newblock Test-time style shifting: Handling arbitrary styles in domain generalization.
\newblock In \emph{Proceedings of the 40th International Conference on Machine Learning (ICML)}, pages 27114--27131, 2023.

\bibitem[Patel et~al.(2016{\natexlab{a}})Patel, Han, Jain, and otehrs.]{2016Secure}
Keyurkumar Patel, Hu Han, Anil~K Jain, and otehrs.
\newblock Secure face unlock: Spoof detection on smartphones.
\newblock \emph{IEEE Transactions on Information Forensics and Security (TIFS)}, 11\penalty0 (10):\penalty0 2268--2283, 2016{\natexlab{a}}.

\bibitem[Patel et~al.(2016{\natexlab{b}})Patel, Han, Jain, et~al.]{DeepBinary02}
Keyurkumar Patel, Hu Han, Anil~K Jain, et~al.
\newblock Cross-database face antispoofing with robust feature representation.
\newblock In \emph{Chinese Conference on Biometric Recognition (CCBR)}, pages 611--619, 2016{\natexlab{b}}.

\bibitem[Qi et~al.(2017)Qi, Yi, Su, and Guibas]{qi2017pointnet++}
Charles~Ruizhongtai Qi, Li Yi, Hao Su, and Leonidas~J Guibas.
\newblock Pointnet++: Deep hierarchical feature learning on point sets in a metric space.
\newblock \emph{Advances in neural information processing systems (NeurIPS)}, 30, 2017.

\bibitem[Qin et~al.(2021)Qin, Yu, Yan, Wang, Zhao, and Lei]{qin2021meta}
Yunxiao Qin, Zitong Yu, Longbin Yan, Zezheng Wang, Chenxu Zhao, and Zhen Lei.
\newblock Meta-teacher for face anti-spoofing.
\newblock \emph{IEEE Transactions on Pattern Analysis and Machine Intelligence (TPAMI)}, 44\penalty0 (10):\penalty0 6311--6326, 2021.

\bibitem[Shao et~al.(2019)Shao, Lan, Li, and Yuen]{shao2019multi}
Rui Shao, Xiangyuan Lan, Jiawei Li, and Pong~C Yuen.
\newblock Multi-adversarial discriminative deep domain generalization for face presentation attack detection.
\newblock In \emph{Proceedings of the IEEE/CVF Conference on Computer Vision and Pattern Recognition (CVPR)}, pages 10023--10031, 2019.

\bibitem[Shao et~al.(2020)Shao, Lan, and Yuen]{shao2020regularized}
Rui Shao, Xiangyuan Lan, and Pong~C Yuen.
\newblock Regularized fine-grained meta face anti-spoofing.
\newblock In \emph{Proceedings of the AAAI Conference on Artificial Intelligence (AAAI)}, pages 11974--11981, 2020.

\bibitem[Song et~al.(2024{\natexlab{a}})Song, Zhou, Li, Fan, Lu, and Ma]{song2024ba}
Yiran Song, Qianyu Zhou, Xiangtai Li, Deng-Ping Fan, Xuequan Lu, and Lizhuang Ma.
\newblock Ba-sam: Scalable bias-mode attention mask for segment anything model.
\newblock In \emph{Proceedings of the IEEE/CVF Conference on Computer Vision and Pattern Recognition (CVPR)}, 2024{\natexlab{a}}.

\bibitem[Song et~al.(2024{\natexlab{b}})Song, Zhou, Lu, Shao, and Ma]{song2024simada}
Yiran Song, Qianyu Zhou, Xuequan Lu, Zhiwen Shao, and Lizhuang Ma.
\newblock Simada: A simple unified framework for adapting segment anything model in underperformed scenes.
\newblock \emph{arXiv preprint arXiv:2401.17803}, 2024{\natexlab{b}}.

\bibitem[Sun et~al.(2023)Sun, Liu, Liu, Li, and Chu]{sun2023rethinking}
Yiyou Sun, Yaojie Liu, Xiaoming Liu, Yixuan Li, and Wen-Sheng Chu.
\newblock Rethinking domain generalization for face anti-spoofing: Separability and alignment.
\newblock In \emph{Proceedings of the IEEE/CVF Conference on Computer Vision and Pattern Recognition (CVPR)}, pages 24563--24574, 2023.

\bibitem[Touvron et~al.(2021)Touvron, Cord, Douze, Massa, Sablayrolles, and J{\'e}gou]{touvron2021training}
Hugo Touvron, Matthieu Cord, Matthijs Douze, Francisco Massa, Alexandre Sablayrolles, and Herv{\'e} J{\'e}gou.
\newblock Training data-efficient image transformers \& distillation through attention.
\newblock In \emph{International Conference on Machine Learning (ICML)}, pages 10347--10357, 2021.

\bibitem[Tzeng et~al.(2017)Tzeng, Hoffman, Saenko, and Darrell]{tzeng2017adversarial}
Eric Tzeng, Judy Hoffman, Kate Saenko, and Trevor Darrell.
\newblock Adversarial discriminative domain adaptation.
\newblock In \emph{Proceedings of the IEEE Conference on Computer Vision and Pattern Recognition (CVPR)}, pages 7167--7176, 2017.

\bibitem[Vallender(1974)]{vallender1974calculation}
SS Vallender.
\newblock Calculation of the wasserstein distance between probability distributions on the line.
\newblock \emph{Theory of Probability \& Its Applications}, 18\penalty0 (4):\penalty0 784--786, 1974.

\bibitem[Van~der Maaten and Hinton(2008)]{van2008visualizing}
Laurens Van~der Maaten and Geoffrey Hinton.
\newblock Visualizing data using t-sne.
\newblock \emph{Journal of Machine Learning Research (JMLR)}, 9\penalty0 (11), 2008.

\bibitem[Wang et~al.(2022{\natexlab{a}})Wang, Lu, Yang, and Lai]{wang2022patchnet}
Chien-Yi Wang, Yu-Ding Lu, Shang-Ta Yang, and Shang-Hong Lai.
\newblock Patchnet: A simple face anti-spoofing framework via fine-grained patch recognition.
\newblock In \emph{Proceedings of the IEEE/CVF Conference on Computer Vision and Pattern Recognition (CVPR)}, pages 20281--20290, 2022{\natexlab{a}}.

\bibitem[Wang et~al.(2020{\natexlab{a}})Wang, Shelhamer, Liu, Olshausen, and Darrell]{wang2020tent}
Dequan Wang, Evan Shelhamer, Shaoteng Liu, Bruno Olshausen, and Trevor Darrell.
\newblock Tent: Fully test-time adaptation by entropy minimization.
\newblock \emph{arXiv preprint arXiv:2006.10726}, 2020{\natexlab{a}}.

\bibitem[Wang et~al.(2021{\natexlab{a}})Wang, Han, Shan, and Chen]{DR-UDA}
Guoqing Wang, Hu Han, Shiguang Shan, and Xilin Chen.
\newblock Unsupervised adversarial domain adaptation for cross-domain face presentation attack detection.
\newblock \emph{IEEE Transactions on Information Forensics and Security (TIFS)}, 16:\penalty0 56--69, 2021{\natexlab{a}}.

\bibitem[Wang et~al.(2021{\natexlab{b}})Wang, Liu, Hu, Shi, and Mei]{wang2021facex}
Jun Wang, Yinglu Liu, Yibo Hu, Hailin Shi, and Tao Mei.
\newblock Facex-zoo: A pytorch toolbox for face recognition.
\newblock In \emph{Proceedings of the 29th ACM International Conference on Multimedia (ACM MM)}, pages 3779--3782, 2021{\natexlab{b}}.

\bibitem[Wang et~al.(2021{\natexlab{c}})Wang, Zhang, Bian, Cai, Wang, and Pu]{wang2021self}
Jingjing Wang, Jingyi Zhang, Ying Bian, Youyi Cai, Chunmao Wang, and Shiliang Pu.
\newblock Self-domain adaptation for face anti-spoofing.
\newblock In \emph{Proceedings of the AAAI conference on artificial intelligence (AAAI)}, pages 2746--2754, 2021{\natexlab{c}}.

\bibitem[Wang et~al.(2021{\natexlab{d}})Wang, Zhao, Jin, Duan, Lei, Huai, Wu, and He]{wang2021vlad}
Jiong Wang, Zhou Zhao, Weike Jin, Xinyu Duan, Zhen Lei, Baoxing Huai, Yiling Wu, and Xiaofei He.
\newblock Vlad-vsa: Cross-domain face presentation attack detection with vocabulary separation and adaptation.
\newblock In \emph{Proceedings of the 29th ACM International Conference on Multimedia (ACM MM)}, pages 1497--1506, 2021{\natexlab{d}}.

\bibitem[Wang et~al.(2022{\natexlab{b}})Wang, Fink, Van~Gool, and Dai]{wang2022continual}
Qin Wang, Olga Fink, Luc Van~Gool, and Dengxin Dai.
\newblock Continual test-time domain adaptation.
\newblock In \emph{Proceedings of the IEEE/CVF Conference on Computer Vision and Pattern Recognition (CVPR)}, pages 7201--7211, 2022{\natexlab{b}}.

\bibitem[Wang et~al.(2021{\natexlab{e}})Wang, Song, Xu, Feng, and Wu]{wang2021rgb}
Yahang Wang, Xiaoning Song, Tianyang Xu, Zhenhua Feng, and Xiao-Jun Wu.
\newblock From rgb to depth: domain transfer network for face anti-spoofing.
\newblock \emph{IEEE Transactions on Information Forensics and Security (TIFS)}, 16:\penalty0 4280--4290, 2021{\natexlab{e}}.

\bibitem[Wang et~al.(2020{\natexlab{b}})Wang, Yu, Zhao, Zhu, Qin, Zhou, Zhou, and Lei]{wang2020deep}
Zezheng Wang, Zitong Yu, Chenxu Zhao, Xiangyu Zhu, Yunxiao Qin, Qiusheng Zhou, Feng Zhou, and Zhen Lei.
\newblock Deep spatial gradient and temporal depth learning for face anti-spoofing.
\newblock In \emph{Proceedings of the IEEE/CVF Conference on Computer Vision and Pattern Recognition (CVPR)}, pages 5042--5051, 2020{\natexlab{b}}.

\bibitem[Wang et~al.(2022{\natexlab{c}})Wang, Wang, Deng, and Guo]{wang2022face}
Zhuo Wang, Qiangchang Wang, Weihong Deng, and Guodong Guo.
\newblock Face anti-spoofing using transformers with relation-aware mechanism.
\newblock \emph{IEEE Transactions on Biometrics, Behavior, and Identity Science (TBIOM)}, 4\penalty0 (3):\penalty0 439--450, 2022{\natexlab{c}}.

\bibitem[Wang et~al.(2022{\natexlab{d}})Wang, Wang, Deng, and Guo]{wang2022learning}
Zhuo Wang, Qiangchang Wang, Weihong Deng, and Guodong Guo.
\newblock Learning multi-granularity temporal characteristics for face anti-spoofing.
\newblock \emph{IEEE Transactions on Information Forensics and Security (TIFS)}, 17:\penalty0 1254--1269, 2022{\natexlab{d}}.

\bibitem[Wang et~al.(2022{\natexlab{e}})Wang, Wang, Yu, Deng, Li, Gao, and Wang]{wang2022domain}
Zhuo Wang, Zezheng Wang, Zitong Yu, Weihong Deng, Jiahong Li, Tingting Gao, and Zhongyuan Wang.
\newblock Domain generalization via shuffled style assembly for face anti-spoofing.
\newblock In \emph{Proceedings of the IEEE/CVF Conference on Computer Vision and Pattern Recognition (CVPR)}, pages 4123--4133, 2022{\natexlab{e}}.

\bibitem[Wang et~al.(2023)Wang, Yu, Wang, Qin, Li, Zhao, Liu, and Lei]{wang2023consistency}
Zezheng Wang, Zitong Yu, Xun Wang, Yunxiao Qin, Jiahong Li, Chenxu Zhao, Xin Liu, and Zhen Lei.
\newblock Consistency regularization for deep face anti-spoofing.
\newblock \emph{IEEE Transactions on Information Forensics and Security (TIFS)}, 18:\penalty0 1127--1140, 2023.

\bibitem[Wen et~al.(2015)Wen, Han, Jain, et~al.]{2015Face}
Di Wen, Hu Han, Anil~K Jain, et~al.
\newblock Face spoof detection with image distortion analysis.
\newblock \emph{IEEE Transactions on Information Forensics and Securityn (TIFS)}, 10\penalty0 (4):\penalty0 746--761, 2015.

\bibitem[Wu et~al.(2021)Wu, Zeng, Hu, Shi, and Mei]{wu2021dual}
Hangtong Wu, Dan Zeng, Yibo Hu, Hailin Shi, and Tao Mei.
\newblock Dual spoof disentanglement generation for face anti-spoofing with depth uncertainty learning.
\newblock \emph{IEEE Transactions on Circuits and Systems for Video Technology (TCSVT)}, 32\penalty0 (7):\penalty0 4626--4638, 2021.

\bibitem[Wu et~al.(2020)Wu, Zhou, Liu, Ni, and Fan]{wu2020single}
Xiaojun Wu, Jinghui Zhou, Jun Liu, Fangyi Ni, and Haoqiang Fan.
\newblock Single-shot face anti-spoofing for dual pixel camera.
\newblock \emph{IEEE Transactions on Information Forensics and Security (TIFS)}, 16:\penalty0 1440--1451, 2020.

\bibitem[Xu et~al.(2021)Xu, Liu, Zhou, Hao, Cao, Feng, and Ma]{xu2021semi}
Hongyi Xu, Fengqi Liu, Qianyu Zhou, Jinkun Hao, Zhijie Cao, Zhengyang Feng, and Lizhuang Ma.
\newblock Semi-supervised 3d object detection via adaptive pseudo-labeling.
\newblock In \emph{IEEE International Conference on Image Processing (ICIP)}, pages 3183--3187, 2021.

\bibitem[Yang et~al.(2013)Yang, Lei, Liao, and Li]{HoG01}
Jianwei Yang, Zhen Lei, Shengcai Liao, and Stan~Z Li.
\newblock Face liveness detection with component dependent descriptor.
\newblock In \emph{IEEE International Conference on Biometrics (ICB)}, pages 1--6, 2013.

\bibitem[Yang et~al.(2014)Yang, Lei, Li, et~al.]{2014Learn}
Jianwei Yang, Zhen Lei, Stan~Z Li, et~al.
\newblock Learn convolutional neural network for face anti-spoofing.
\newblock In \emph{arXiv preprint arXiv:1408.5601}, 2014.

\bibitem[Yu et~al.(2020{\natexlab{a}})Yu, Li, Niu, Shi, and Zhao]{BCN}
Zitong Yu, Xiaobai Li, Xuesong Niu, Jingang Shi, and Guoying Zhao.
\newblock Face anti-spoofing with human material perception.
\newblock In \emph{European Conference on Computer Vision (ECCV)}, pages 557--575, 2020{\natexlab{a}}.

\bibitem[Yu et~al.(2020{\natexlab{b}})Yu, Wan, Qin, Li, Li, and Zhao]{yu2020fas}
Zitong Yu, Jun Wan, Yunxiao Qin, Xiaobai Li, Stan~Z Li, and Guoying Zhao.
\newblock Nas-fas: Static-dynamic central difference network search for face anti-spoofing.
\newblock \emph{IEEE Transactions on Pattern Analysis and Machine Intelligence (TPAMI)}, 43\penalty0 (9):\penalty0 3005--3023, 2020{\natexlab{b}}.

\bibitem[Yu et~al.(2020{\natexlab{c}})Yu, Zhao, Wang, Qin, Su, Li, Zhou, and Zhao]{CDCN}
Zitong Yu, Chenxu Zhao, Zezheng Wang, Yunxiao Qin, Zhuo Su, Xiaobai Li, Feng Zhou, and Guoying Zhao.
\newblock Searching central difference convolutional networks for face anti-spoofing.
\newblock In \emph{Proceedings of the IEEE/CVF Conference on Computer Vision and Pattern Recognition (CVPR)}, pages 5295--5305, 2020{\natexlab{c}}.

\bibitem[Yu et~al.(2021{\natexlab{a}})Yu, Li, Shi, Xia, and Zhao]{yu2021revisiting}
Zitong Yu, Xiaobai Li, Jingang Shi, Zhaoqiang Xia, and Guoying Zhao.
\newblock Revisiting pixel-wise supervision for face anti-spoofing.
\newblock \emph{IEEE Transactions on Biometrics, Behavior, and Identity Science (TBIOM)}, 3\penalty0 (3):\penalty0 285--295, 2021{\natexlab{a}}.

\bibitem[Yu et~al.(2021{\natexlab{b}})Yu, Qin, Zhao, Li, and Zhao]{yu2021dual}
Zitong Yu, Yunxiao Qin, Hengshuang Zhao, Xiaobai Li, and Guoying Zhao.
\newblock Dual-cross central difference network for face anti-spoofing.
\newblock In \emph{Proceedings of the Thirtieth International Joint Conference on Artificial Intelligence (IJCAI)}, pages 1281--1287, 2021{\natexlab{b}}.

\bibitem[Zhang et~al.(2020)Zhang, Yao, Zhang, Tai, Ding, Li, Huang, Song, and Ma]{disentangle01}
Ke-Yue Zhang, Taiping Yao, Jian Zhang, Ying Tai, Shouhong Ding, Jilin Li, Feiyue Huang, Haichuan Song, and Lizhuang Ma.
\newblock Face anti-spoofing via disentangled representation learning.
\newblock In \emph{European Conference on Computer Vision (ECCV)}, pages 641--657, 2020.

\bibitem[Zhang et~al.(2021)Zhang, Yao, Zhang, Liu, Yin, Ding, and Li]{zhang2021structure}
Ke-Yue Zhang, Taiping Yao, Jian Zhang, Shice Liu, Bangjie Yin, Shouhong Ding, and Jilin Li.
\newblock Structure destruction and content combination for face anti-spoofing.
\newblock In \emph{IEEE International Joint Conference on Biometrics (IJCB)}, pages 1--6, 2021.

\bibitem[Zhang et~al.(2022)Zhang, Levine, and Finn]{zhang2022memo}
Marvin Zhang, Sergey Levine, and Chelsea Finn.
\newblock Memo: Test time robustness via adaptation and augmentation.
\newblock \emph{Advances in Neural Information Processing Systems (NeurIPS)}, 35:\penalty0 38629--38642, 2022.

\bibitem[Zhang et~al.(2012)Zhang, Yan, Liu, Lei, Yi, and Li]{Zhang2012A}
Zhiwei Zhang, Junjie Yan, Sifei Liu, Zhen Lei, Dong Yi, and Stan~Z Li.
\newblock A face antispoofing database with diverse attacks.
\newblock In \emph{5th IAPR International Conference on Biometrics (ICB)}, pages 26--31, 2012.

\bibitem[Zhao et~al.(2022)Zhao, Liu, Sicilia, Hwang, and Fu]{ijcai2022p240}
Xingchen Zhao, Chang Liu, Anthony Sicilia, Seong~Jae Hwang, and Yun Fu.
\newblock Test-time fourier style calibration for domain generalization.
\newblock In \emph{Proceedings of the Thirty-First International Joint Conference on Artificial Intelligence (IJCAI)}, pages 1721--1727, 2022.

\bibitem[Zheng et~al.(2023)Zheng, Liu, Dai, Li, Zou, and Xiong]{zheng2023learning}
Guanghao Zheng, Yuchen Liu, Wenrui Dai, Chenglin Li, Junni Zou, and Hongkai Xiong.
\newblock Learning causal representations for generalizable face anti spoofing.
\newblock In \emph{IEEE International Conference on Acoustics, Speech and Signal Processing (ICASSP)}, pages 1--5, 2023.

\bibitem[Zhou et~al.(2022{\natexlab{a}})Zhou, Feng, Gu, Cheng, Lu, Shi, and Ma]{zhou2020uncertainty}
Qianyu Zhou, Zhengyang Feng, Qiqi Gu, Guangliang Cheng, Xuequan Lu, Jianping Shi, and Lizhuang Ma.
\newblock Uncertainty-aware consistency regularization for cross-domain semantic segmentation.
\newblock \emph{Computer Vision and Image Understanding (CVIU)}, page 103448, 2022{\natexlab{a}}.

\bibitem[Zhou et~al.(2022{\natexlab{b}})Zhou, Zhang, Yao, Yi, Ding, and Ma]{zhou2022adaptive}
Qianyu Zhou, Ke-Yue Zhang, Taiping Yao, Ran Yi, Shouhong Ding, and Lizhuang Ma.
\newblock Adaptive mixture of experts learning for generalizable face anti-spoofing.
\newblock In \emph{Proceedings of the 30th ACM International Conference on Multimedia (ACM MM)}, pages 6009--6018, 2022{\natexlab{b}}.

\bibitem[Zhou et~al.(2022{\natexlab{c}})Zhou, Zhang, Yao, Yi, Sheng, Ding, and Ma]{zhou2022generative}
Qianyu Zhou, Ke-Yue Zhang, Taiping Yao, Ran Yi, Kekai Sheng, Shouhong Ding, and Lizhuang Ma.
\newblock Generative domain adaptation for face anti-spoofing.
\newblock In \emph{European Conference on Computer Vision (ECCV)}, pages 335--356, 2022{\natexlab{c}}.

\bibitem[Zhou et~al.(2022{\natexlab{d}})Zhou, Zhuang, Lu, and Ma]{zhou2022domain}
Qianyu Zhou, Chuyun Zhuang, Xuequan Lu, and Lizhuang Ma.
\newblock Domain adaptive semantic segmentation with regional contrastive consistency regularization.
\newblock In \emph{IEEE International Conference on Multimedia and Expo (ICME)}, 2022{\natexlab{d}}.

\bibitem[Zhou et~al.(2023{\natexlab{a}})Zhou, Feng, Gu, Pang, Cheng, Lu, Shi, and Ma]{zhou2022context}
Qianyu Zhou, Zhengyang Feng, Qiqi Gu, Jiangmiao Pang, Guangliang Cheng, Xuequan Lu, Jianping Shi, and Lizhuang Ma.
\newblock Context-aware mixup for domain adaptive semantic segmentation.
\newblock \emph{IEEE Transactions on Circuits and Systems for Video Technology (TCSVT)}, 33\penalty0 (2):\penalty0 804--817, 2023{\natexlab{a}}.

\bibitem[Zhou et~al.(2023{\natexlab{b}})Zhou, Gu, Pang, Lu, and Ma]{zhou2023self}
Qianyu Zhou, Qiqi Gu, Jiangmiao Pang, Xuequan Lu, and Lizhuang Ma.
\newblock Self-adversarial disentangling for specific domain adaptation.
\newblock \emph{IEEE Transactions on Pattern Analysis and Machine Intelligence (TPAMI)}, 45\penalty0 (7):\penalty0 8954--8968, 2023{\natexlab{b}}.

\bibitem[Zhou et~al.(2023{\natexlab{c}})Zhou, Zhang, Yao, Lu, Yi, Ding, and Ma]{zhou2023instance}
Qianyu Zhou, Ke-Yue Zhang, Taiping Yao, Xuequan Lu, Ran Yi, Shouhong Ding, and Lizhuang Ma.
\newblock Instance-aware domain generalization for face anti-spoofing.
\newblock In \emph{Proceedings of the IEEE/CVF Conference on Computer Vision and Pattern Recognition (CVPR)}, pages 20453--20463, 2023{\natexlab{c}}.

\end{thebibliography}
}
%

%
%
%
%

\end{document}